\documentclass[11pt]{article}

\usepackage[final]{acl}

\usepackage{times}
\usepackage{latexsym}

\usepackage[T1]{fontenc}

\usepackage[utf8]{inputenc}

\usepackage{microtype}

\usepackage{inconsolata}

\usepackage{algorithm}
\usepackage{algorithmic}

\usepackage{amsfonts}
\usepackage{amssymb}

\usepackage{bm}
\usepackage{multirow}
\usepackage{multicol}

\usepackage{graphicx}
\usepackage{booktabs}
\usepackage{framed}
\usepackage{caption}
\usepackage{subcaption}
\usepackage{xspace}
\usepackage{enumitem}
\usepackage{xcolor}
\usepackage{color, colortbl}
\usepackage{amsthm}
\usepackage{amsmath}
\usepackage{tcolorbox}
\usepackage{cleveref}
\usepackage[inkscapelatex=false]{svg}
\usepackage{dashrule}
\usepackage{arydshln}
\usepackage{overpic}
\usepackage{marvosym}
\usepackage{adjustbox}

\def\strong{\textit{Strong}\xspace}
\def\mix{\textit{Mix}\xspace}
\def\car{\textit{CAR}\xspace}
\def\family{\textit{Family-Strong}\xspace}

\newcommand{\cdashlinelr}[1]{%
  \noalign{\vskip\aboverulesep
           \global\let\@dashdrawstore\adl@draw
           \global\let\adl@draw\adl@drawiv}
  \cdashline{#1}
  \noalign{\global\let\adl@draw\@dashdrawstore
           \vskip\belowrulesep}}

\makeatletter
\def\@fnsymbol#1{}
\makeatother

\newtheorem{insight}{Finding}

\newenvironment{tcolorboxinsight}[1][]{%
    \begin{tcolorbox}[#1]
    \vspace{-0.2cm}
    \begin{insight}
}{%
    \end{insight}
    \vspace{-0.3cm}
    \end{tcolorbox}
}

\title{Find Your Optimal Teacher: Personalized Data Synthesis via Router-Guided Multi-Teacher Distillation}

\author{
 \textbf{Hengyuan Zhang\textsuperscript{1 *}},
 \textbf{Shiping Yang\textsuperscript{2 *}}\thanks{*\ Equal contribution.},
 \textbf{Xiao Liang\textsuperscript{3}},
 \textbf{Chenming Shang\textsuperscript{4}},
 \textbf{Yuxuan Jiang\textsuperscript{5}},\\
 \textbf{Chaofan Tao\textsuperscript{1}}, 
 \textbf{Jing Xiong\textsuperscript{1}},
 \textbf{Hayden Kwok-Hay So\textsuperscript{1}}, 
 \textbf{Ruobing Xie\textsuperscript{6}},
 \textbf{Angel X. Chang\textsuperscript{2}},
 \textbf{Ngai Wong\textsuperscript{1 \dag}}\thanks{\dag\ Corresponding author.}
\\
 \textsuperscript{1}The University of Hong Kong  \ \
 \textsuperscript{2}Simon Fraser University  \ \ 
 \textsuperscript{3}University of California, Los Angeles  \ \ \\
 \textsuperscript{4}Dartmouth College \ \
 \textsuperscript{5}University of Maryland, Baltimore County  \ \
 \textsuperscript{6}Tencent 
 \\
\texttt{hengyuan.zhang88@gmail.com }  
}

\begin{document}
\maketitle

\begin{abstract}
Training student models on synthetic data generated by strong teacher models is a promising way to distilling the capabilities of teachers.
However, recent studies show that stronger models are not always optimal teachers, revealing a mismatch between teacher outputs and student learnability.
To address this issue, we propose \textit{PerSyn} (\underline{\textbf{Per}}sonalized data \underline{\textbf{Syn}}thesis), a novel synthesis strategy that operates under a new ``Route then Generate'' paradigm to create data tailored to each student model, enabling it to learn more effectively.
Specifically, \textit{PerSyn} first assigns each prompt to its optimal teacher via a query-level router that jointly considers student learnability and teacher response quality. 
Each teacher then synthesizes data only for its assigned prompts, making the process more efficient than the conventional ``Generate then Select'' paradigm, where all teachers must generate parallel responses for the entire prompt set before constructing the final dataset.
Extensive experiments across different model families and scales demonstrate that \textit{PerSyn} consistently achieves superior or comparable performance to all baselines in instruct tuning and math reasoning settings.
Further analysis verifies the effectiveness of \textit{PerSyn} and offers extra insights to propel future research.
The code is available at \href{https://github.com/rattlesnakey/PerSyn}{the provided link}.
\end{abstract}

\section{Introduction}
\label{sec:intro}
Large Language Models (LLMs) have demonstrated outstanding performance across a wide range of applications, such as reasoning~\citep{li2025system,ren2025deepseek,yu2025chain,liang2026training}, multilingualism~\citep{gurgurov2024multilingual,zhang2024shifcon,qin2025survey}, and other specialized domains~\citep{yang2024llm,zhang2024balancing,zhao2024revolutionizing,chang2025treereview}.
However, the high computational cost of LLMs hinders their deployment on resource-constrained devices, motivating the development of smaller models that offer similar capabilities at reduced cost.
A common strategy to achieve this is distillation~\citep{hinton2015distillingknowledgeneuralnetwork,kim2024promptkddistillingstudentfriendlyknowledge,wang2025distilling}, which leverages the synthetic data generated by a strong teacher model to fine-tune a small student model.
They assume that stronger teacher will produce higher-quality synthetic data, which in turn enables the student model to learn more effectively.

\begin{table}[!t]
\centering
\setlength\tabcolsep{5pt}
\fontsize{9}{11}\selectfont 
\begin{tabular}{lcccc}
\toprule[1.5pt]
                & Strong & Mix & CAR & PerSyn \\ \addlinespace[2pt] \hline \addlinespace[2pt]
Quality & \adjustbox{valign=b}{\includegraphics{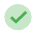}}    & 
\adjustbox{valign=b}{\includegraphics[width=0.03\textwidth]{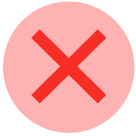}}                     &
\adjustbox{valign=b}{\includegraphics{figs/correct.pdf}}         & \adjustbox{valign=b}{\includegraphics{figs/correct.pdf}}                                                        \\
Learnability  & \adjustbox{valign=b}{\includegraphics[width=0.03\textwidth]{figs/wrong.pdf}}    & 
\adjustbox{valign=b}{\includegraphics{figs/correct.pdf}} & 
\adjustbox{valign=b}{\includegraphics{figs/correct.pdf}}         & \adjustbox{valign=b}{\includegraphics{figs/correct.pdf}}                                                        \\
Efficiency  & \adjustbox{valign=b}{\includegraphics[width=0.03\textwidth]{figs/wrong.pdf}} & \adjustbox{valign=b}{\includegraphics[width=0.03\textwidth]{figs/wrong.pdf}}  & 
\adjustbox{valign=b}{\includegraphics[width=0.03\textwidth]{figs/wrong.pdf}}  & \adjustbox{valign=b, raise=-1pt}{\includegraphics{figs/correct.pdf}}                              \\
Sample Level  & \adjustbox{valign=b}{\includegraphics[width=0.03\textwidth]{figs/wrong.pdf}}  & \adjustbox{valign=b}{\includegraphics[width=0.03\textwidth]{figs/wrong.pdf}}   & \adjustbox{valign=b}{\includegraphics[width=0.03\textwidth]{figs/wrong.pdf}}         & \adjustbox{valign=b, raise=-1pt}{\includegraphics{figs/correct.pdf}}                              \\\bottomrule[1.5pt]
\end{tabular}
\caption{Compared to existing methods, \textit{PerSyn} can efficiently assigns each prompt to the optimal teacher by jointly considering both teacher quality and student learnability. ``Sample Level'' indicates whether each prompt is assigned to the optimal teacher. \strong uses the strongest model as teacher, \mix combines synthetic data from strong and weak teachers, and \car selects a single teacher balancing quality and compatibility.}
\label{tab:intro}
\vspace{-0.4cm}
\end{table}

Nevertheless, some works~\citep{stronger_are_not,learnability_gap} demonstrate that stronger models are not always the optimal teachers for small student models, since their outputs may be overly complex and shift away from the students' distribution.
To mitigate this issue, \citet{learnability_gap} mixed the synthetic data from strong and weak models (\textit{Mix}).
\citet{stronger_are_not} designed a Compatibility-Adjusted Reward (\textit{CAR}) metric to select a single appropriate teacher model from a pool of teacher models for specific student.
Despite these efforts, two critical limitations remain, as shown in Table~\ref{tab:intro}: \textbf{1)} These methods are not efficient enough. 
Specifically, \strong refers to using a super-sized LLM for distillation, but often yields sub-optimal performance with high computational cost.
\mix and \car follow the ``Generate then Select'' paradigm, which requires parallel teacher responses\footnote{In this paper, parallel teacher responses denote the responses generated by all teacher models for a given prompt.}, thereby all candidate teacher models must generate responses for the entire prompt set before constructing the final synthetic dataset. Notably, the cost scales linearly with the teacher model pool size; \textbf{2)} These methods also overlook that each prompt within the dataset has its corresponding optimal teacher for synthesizing responses, thereby making the synthetic dataset sub-optimal for student.

To address these limitations, we propose \textit{PerSyn} (\underline{\textbf{Per}}sonalized data \underline{\textbf{Syn}}thesis), a novel synthesis strategy that customizes a synthetic dataset for a specific student model to help it learn more effectively.
Specifically, unlike the ``Generate then Select'' paradigm, our method operates in a more efficient manner, i.e, ``Route then Generate'', which first assigns each prompt to its corresponding optimal teacher model, and then the teacher only needs to synthesize the assigned prompts. 
The assigning process is achieved by a router-guided mechanism with considering both the student model's learnability and teacher model's response quality.
Moreover, further analysis reveals that over 95\% of prompts are routed to smaller teacher models (unlike the \strong baseline, which relies on a single super-sized LLM for all prompts), leading to more efficient synthesis.

To summarize, our contributions are as follows:

\vspace{0.3em}
\quad \textbf{1)} To construct personalized synthetic dataset for specific student model, we propose \textit{PerSyn}, an efficient strategy that transfers the synthesis paradigm from ``Generate then Select'' to a more efficient manner ``Route then Generate''. In this paradigm, each prompt is first routed to its optimal teacher based on both learnability and quality, and each teacher model is then responsible only for synthesizing the prompts assigned to it.

\vspace{0.3em}
\quad \textbf{2)} Extensive experiments validate the effectiveness of \textit{PerSyn} across different model families and scales in two common distillation settings (e.g., 8.7\% on IFEval, and 7.5\% on MATH). We also construct a math synthetic dataset \textbf{\textit{PerSyn-Math}}, which includes parallel responses from 15 teacher models to facilitate future research.

\vspace{0.3em}
\quad \textbf{3)} Further analysis offer valuable insights into the routing behavior of \textit{PerSyn}. For example, both quality and learnability are important in \textit{PerSyn}, with quality playing a more critical role.

\section{PerSyn}
\label{sec:method}
\begin{figure*}[!th]
    \centering \centerline{\includegraphics[width=2\columnwidth]{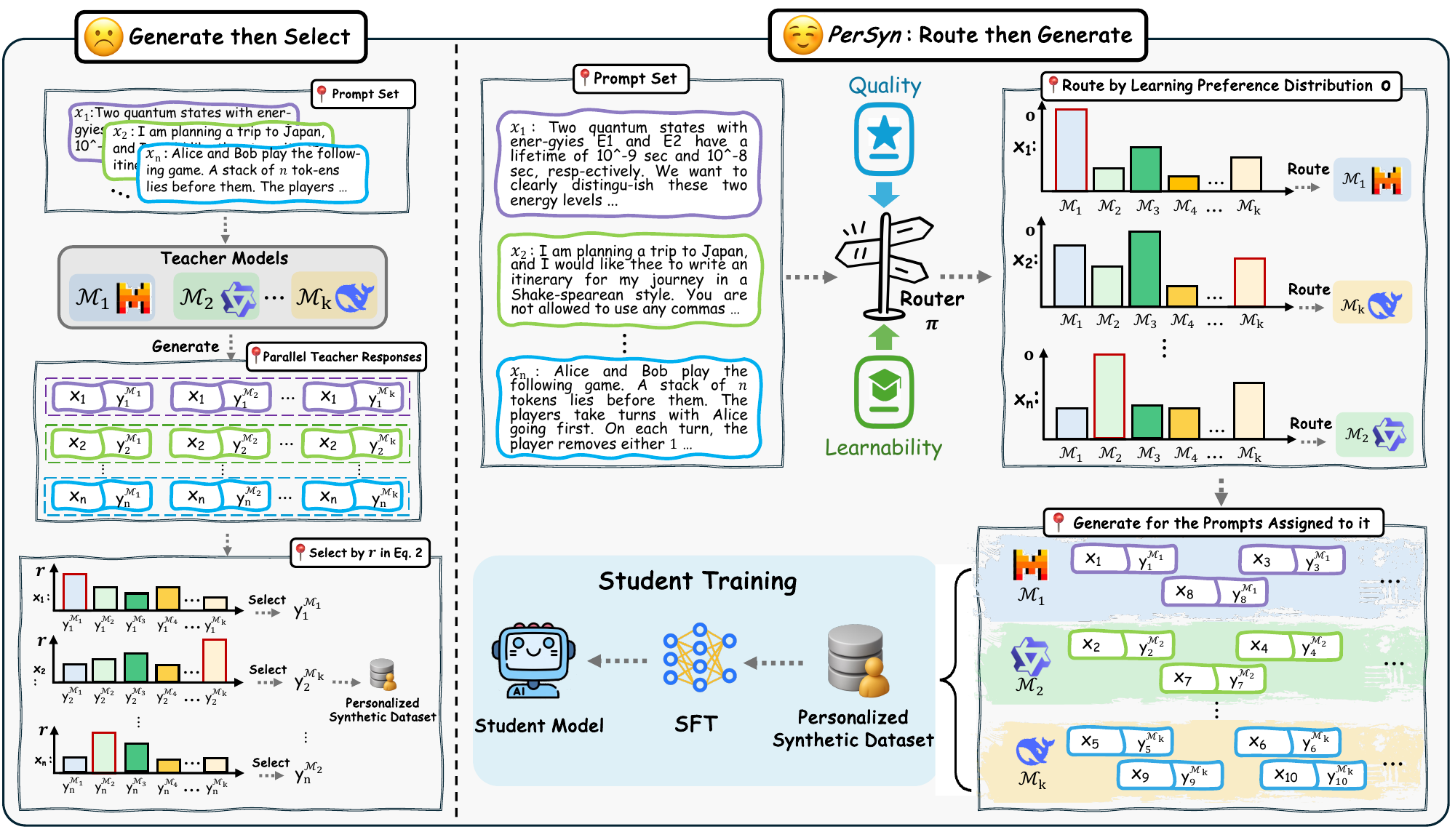}}
    \vspace{-0.1cm}
    \caption{Overview of the two paradigm for obtaining a personalized synthetic dataset. The left part illustrates how we select optimal teacher response for each prompt using the proposed criterion. This process follows the conventional ``Generate then Select'' approach, which requires parallel teacher responses for the entire prompt set (\S\ref{sec:annotation_metric}). 
In contrast, \textit{PerSyn} adopts a more efficient ``Route then Generate'' paradigm: it first routes each prompt to an optimal teacher based on learnability and quality via a router-guided mechanism, and teachers generate responses only for their assigned prompts (\S\ref{sec:transfer_paradigm}). Details of router training are described in \S\ref{sec:obtain_router}.}
\vspace{-0.2cm}
    \label{fig:overview}
\end{figure*}

In this section, we first present the criterion used by \textit{PerSyn} to find the optimal teacher model for each prompt (\S\ref{sec:annotation_metric}). 
Next, we describe how \textit{PerSyn} transfers the ``Generate then Select'' paradigm to the more efficient ``Route then Generate'' paradigm with a router, and how the resulting synthetic data is used to train the student model (\S\ref{sec:transfer_paradigm}). 
Finally, we illustrate how the \textit{PerSyn} router is obtained (\S\ref{sec:obtain_router}). 
Fig.~\ref{fig:overview} shows the overview of \textit{PerSyn} strategy.

\subsection{Finding the Optimal Teacher}
\label{sec:annotation_metric}
Given a prompt $x_i$, a straightforward way to select the optimal teacher's response for a student model $\theta$ is the ``Generate then Select'' paradigm. 
This approach first lets all teacher models $\mathcal{M} = \{\mathcal{M}_1, \mathcal{M}_2, \dots, \mathcal{M}_n\}$ generate parallel responses $\mathcal{Y}_i = \{y^{\mathcal{M}_1}_i, y^{\mathcal{M}_2}_i, \dots, y^{\mathcal{M}_n}_i\}$ for $x_i$ (where $y^{\mathcal{M}_n}_i$ is the response from $\mathcal{M}_n$), and then selects the optimal response for the student model $\theta$ from $\mathcal{Y}_i$.

To identify the optimal response, we evaluate each $y^{\mathcal{M}_n}_i$ using two complementary reward.
The first is the learnability reward, which measures how easily the student model $\theta$ can learn from $y^{\mathcal{M}_n}_i$. 
Responses that are too difficult for $\theta$, i.e., have a large learnability gap, tend to make learning inefficient~\citep{learnability_gap,stronger_are_not}. We compute the learnability reward $\boldsymbol{r_l}(y^{\mathcal{M}_n}_i, \theta)$ using the student’s self-derived log-likelihood:
\begin{equation}
\small
\boldsymbol{r_l}(y^{\mathcal{M}_n}_i, \theta) = \frac{1}{|y^{\mathcal{M}_n}_i|}\sum_{t=1}^{|y^{\mathcal{M}_n}_i|} \log p_{\pi}\bigl(y^{\mathcal{M}_n (t)}_{i} \bigm| y^{\mathcal{M}_n(<t)}_i, x_i\bigr),
\label{eq:learnability_score}
\end{equation}
\noindent where $\log p_{\pi}\bigl(y^{\mathcal{M}_i (t)}_n \bigm| y^{\mathcal{M}_n(<t)}_i, x_i\bigr)$ is the probability assigned by the student to the $t$-th token of $y^{\mathcal{M}_n}_i$, given its preceding tokens and the prompt $x_i$. Intuitively, a higher $\boldsymbol{r_l}(y^{\mathcal{M}_n}_i, \theta)$ indicates that $y^{\mathcal{M}_n}_i$ aligns well with the student’s existing knowledge and capabilities, therefore is more learnable.

However, learnability alone is insufficient. For instance, a response may be highly learnable yet trivial or low-quality, offering little benefit to the student. To account for this, we introduce a quality reward $\boldsymbol{r_q}(y^{\mathcal{M}_n}_i)$, estimated by a reward model, where larger values indicate higher quality.

The overall reward of $y^{\mathcal{M}_n}_i$ for $\theta$ is then computed as a weighted combination of two aspects:
\vspace{-1em}
\begin{equation}
\boldsymbol{r}(y^{\mathcal{M}_n}_i, \theta) = (1 - \alpha) \boldsymbol{r_q}(y^{\mathcal{M}_n}_i) + \alpha \boldsymbol{r_l}(y^{\mathcal{M}_n}_i, \theta),
\label{eq:overall_score}
\end{equation}

\noindent where $\alpha$ balances the contribution of learnability and quality rewards.\footnote{Both rewards are normalized across teacher models before combination.}
Finally, the teacher model corresponding to the response with the highest $\boldsymbol{r}(y_n, \theta)$ is selected as optimal.

\subsection{Transfer Paradigm via \textit{PerSyn} Router}
\label{sec:transfer_paradigm}

However, the ``Generate then Select'' paradigm is inefficient. 
For example, synthesizing a dataset of 100K prompts with 20 teacher models would require 2,000K total generations, which is costly and impractical. 
To address this, we introduce \textit{PerSyn} router, which transfers the paradigm to a more efficient ``Route then Generate'' manner: each prompt is first routed to its optimal teacher, and each teacher generates responses only for the prompts assigned to it (Fig.~\ref{fig:overview} illustrates the comparison between the two paradigms).  

More specifically, under the paradigm of ``Route then Generate'', suppose we have a prompt set $\mathcal{X}=\{x_1, x_2, \dots, x_{10}\}$ and a teacher model set $\mathcal{M} = \{\mathcal{M}_1, \mathcal{M}_2, \mathcal{M}_3\}$, the \textit{PerSyn} router will first assigns each prompt to its optimal teacher. 
For example, $\mathcal{M}_1$ is assigned $\mathcal{X}_{\mathcal{M}_1} = \{x_1, x_3, x_8\}$, $\mathcal{M}_2$ is assigned $\mathcal{X}_{\mathcal{M}_2} = \{x_2, x_4, x_7\}$, and $\mathcal{M}_3$ is assigned $\mathcal{X}_{\mathcal{M}_3} = \{x_5, x_6, x_9, x_{10}\}$. 
Each teacher $\mathcal{M}_i$ then generates responses only for its assigned subset $\mathcal{X}_{\mathcal{M}_i}$ to obtain $\mathcal{D}_{\mathcal{M}_i}$ (e.g., $\mathcal{D}_{\mathcal{M}_2} = \{(x_2, y^{\mathcal{M}_2}_2), (x_4, y^{\mathcal{M}_2}_4), (x_7, y^{\mathcal{M}_2}_7)\}$). The final synthetic dataset is $\mathcal{D} = \{\mathcal{D}_{\mathcal{M_\text{i}}}\}_{i=1}^{|\mathcal{M}|}$.  
The student model is then trained via supervised fine-tuning (SFT) using the synthetic dataset $\mathcal{D}$, where the loss is computed only on the response tokens.

\subsection{Obtaining \textit{PerSyn} Router}
\label{sec:obtain_router}
Formally, given a prompt $x$, the \textit{PerSyn} router $\pi$ outputs a score vector $\mathbf{o} = \pi(x) \in \mathbb{R}^{|\mathcal{M}|}$, which reflects the student model $\theta$'s learning preferences over the teacher model set $\mathcal{M}$. 
Each component $\mathbf{o}_i$ represents the latent preference score (i.e., logit) assigned to teacher model $\mathcal{M}_i$ for prompt $x$. 
Importantly, $\mathbf{o}$ is not a normalized probability distribution; rather, it is an unnormalized score vector whose values vary across prompts, capturing the intuition that different teachers may excel at different queries.

To model the student's learning preferences, we adopt the Bradley-Terry (BT) model~\citep{bradley1952rank}. 
Given a comparison between two teacher models $A$ and $B$, we first consider their latent preference scores produced by the router. 
To decide which teacher is preferred, we need to convert the difference between these two scores into a probability $P(B \succ A \mid x)$. 
The BT model provides a natural way to do this by defining the pairwise preference probability as the sigmoid of the score difference, which maps it into the $[0,1]$ range. 
Formally, the probability that $B$ is preferred over $A$ is defined as:
\begin{equation}
\label{eq:BT}
    \mathbb{P}(C = B \succ A \mid Z = z, X = x) = \sigma(z^\top \pi(x)),
\end{equation}
\noindent where $z$ is a `two-hot' encoding of the model comparison pair $(A, B)$, i.e., a vector of length $|\mathcal{M}|$ with $+1$ at the index of $B$, $-1$ at the index of $A$, and zeros elsewhere. 
Here, $\sigma$ denotes the sigmoid function. 
The label $C = 1$ indicates that $B$ is preferred over $A$, and $C = 0$ otherwise. 
Thus, $\sigma(z^\top \pi(x))$ gives the probability of model $B$ being favored over model $A$ for student $\pi$ on prompt $x$.

To construct the pairwise preference dataset $\mathcal{K} = \{(X, Z, C)\}_{i=1}^N$, we first sample a small subset of prompts $\mathcal{X}_{\text{sub}} \subset \mathcal{X}$, and query all teacher models $\mathcal{M}$ to obtain their parallel responses, forming $\mathcal{P}_{\text{sub}}$. 
Each response is then scored using the reward metric defined in Eq.~\ref{eq:overall_score} (see \S\ref{sec:annotation_metric}), which allows us to derive exact teacher rankings and subsequently generate pairwise comparisons (See Appendix~\ref{appendix:details_of_router} for more details about the pairwise dataset construction).

Finally, given pairwise learning preference dataset $\mathcal{K}$, the \textit{PerSyn} router is learned by minimizing:
\begin{equation}
   \hat{\pi} = {\text{argmin}} \frac{1}{N} \sum_{i=1}^{N} \mathcal{L}(\sigma(Z_i^{\top} \pi(X_i)), C_i).
\end{equation}
\noindent where $\mathcal{L}$ denotes the binary cross-entropy loss computed between the predicted pairwise preference probability $\sigma(Z_i^{\top} \pi(X_i))$ and the ground-truth label $C_i$.
By default, we instantiate the \textit{PerSyn} router on top of Qwen2.5-1.5B, as supported by the experiments in \S\ref{par:router_performance}. Specifically, we remove the original language modeling head and replace it with a new coefficient head. In the BT setting, this coefficient head is implemented as a linear layer whose output dimension equals the number of teacher models. 
Note that for each student model, a separate router is needed for each setting.

\section{Experiments}
\label{sec:experiments}
In this section, we conduct experiments on two common distillation settings, instruction tuning and math reasoning, to validate the effectiveness of our \textit{PerSyn} strategy.

\subsection{Experiment Settings}
\label{sec:exp_setting}
\paragraph{Datasets.}
To obtain the synthetic dataset for training the student model in both settings, we proceed as follows.
For instruction tuning, we randomly sample 50K prompts from the Magpie-100K-Generator-Zoo (Magpie-Zoo)\footnote{\url{https://huggingface.co/datasets/Magpie-Align/Magpie-100K-Generator-Zoo}} to construct the training dataset. 
The distilled student models are evaluated on TruthfulQA~\citep{truthfulqa}, LiveBench~\citep{livebench}, and IFEval~\citep{ifeval}.  
For mathematical reasoning, we construct a dataset with parallel teacher responses, denoted \textbf{\textit{PerSyn-Math}}, by randomly sampling 10K queries from OpenR1-Math-220K\footnote{\url{https://huggingface.co/datasets/open-r1/OpenR1-Math-220k}} as seed data and distilling responses from 15 teacher models. 
This dataset is then used to build the synthetic training dataset for student models.\footnote{In the instruction-tuning setting, Magpie-Zoo already provides parallel teacher responses for each prompt.} The resulting student models are evaluated on SVAMP~\citep{svamp}, MATH~\citep{math}, and GSM8K~\citep{gsm8k}.  
Additional details about datasets and evaluation can be found in Appendix~\ref{appendix:dataset_details}. To obtain the \textit{PerSyn} router, we construct parallel teacher responses for 2.5K prompts in both settings to build pairwise preference data.\footnote{As shown in \S\ref{par:router_performance}, 2.5K samples with parallel teacher responses are sufficient to obtain a well-performing \textit{PerSyn} router.}

\begin{table*}[!t]
\centering
\setlength\tabcolsep{4pt}
\fontsize{9.5}{7.5}\selectfont
\begin{tabular}{llccccccc}
\toprule[1.5pt]
\toprule[1pt]
\textbf{Student Model}        & \textbf{Strategy} & \textbf{IFEval} & \textbf{TruthfulQA} & \textbf{LiveBench} & \textbf{GSM8K} & \textbf{MATH}  & \textbf{SVAMP} & \textbf{Avg.}  \\ \addlinespace[2pt]\hline\addlinespace[2pt]
\multirow{5}{*}{Qwen2.5-0.5B} & Strong           & 25.59           & 39.89               & 8.40               & 30.37          & 15.20          & 51.60          & 28.51          \\
                              & Mix           & 26.06           & 40.54               & 8.10               & 33.83          & 20.60          & 55.40          & 30.75          \\
                              & Family-Strong       & 26.75           & 41.43               & 8.60               & 35.62          & 22.80          & 57.00          & 32.03          \\
                              & CAR               & 27.11           & 41.85               & 9.00               & 36.76          & 24.00          & 57.90          & 32.77          \\
                              & PerSyn (Ours)      & \textbf{28.73}  & \textbf{43.01}      & \textbf{9.80}      & \textbf{38.25} & \textbf{25.60} & \textbf{59.40} & \textbf{34.13} \\ \addlinespace[2pt] \hdashline[1pt/1pt] \addlinespace[2pt]
\multirow{5}{*}{Qwen2.5-1.5B} & Strong           & 31.52           & 49.04               & 12.80              & 64.83          & 44.20          & 78.50          & 46.82          \\
                              & Mix           & 31.98           & 49.73               & 13.30              & 65.68          & 45.80          & 80.30          & 47.79          \\
                              & Family-Strong       & 32.63           & 50.45               & 13.60              & 66.55          & 47.40          & 81.20          & 48.64          \\
                              & CAR               & 33.06           & 50.98               & 13.30              & 67.37          & 48.60          & 81.90          & 49.21          \\
                              & PerSyn (Ours)      & \textbf{34.15}  & \textbf{52.22}      & \textbf{14.80}     & \textbf{68.81} & \textbf{50.40} & \textbf{83.40} & \textbf{50.63} \\ \addlinespace[2pt] \hdashline[1pt/1pt] \addlinespace[2pt]
\multirow{5}{*}{Gemma-2-2B}   & Strong           & 28.84           & 40.17               & 10.30              & 29.71          & 14.20          & 47.50          & 28.45          \\
                              & Mix           & 29.39           & 40.83               & 10.90              & 31.66          & 16.40          & 49.40          & 29.76          \\
                              & Family-Strong       & 29.76           & 41.64               & 11.60              & 30.43          & 15.80          & 48.10          & 29.56          \\
                              & CAR               & 30.11           & 42.28               & \textbf{12.80}     & 33.25          & 19.20          & 50.80          & 31.41          \\
                              & PerSyn (Ours)      & \textbf{31.25}  & \textbf{43.87}      & 12.40              & \textbf{35.57} & \textbf{21.40} & \textbf{52.60} & \textbf{32.85} \\ \addlinespace[2pt] \hdashline[1pt/1pt] \addlinespace[2pt]
\multirow{5}{*}{Qwen2.5-3B}   & Strong           & 40.61           & 51.21               & 19.10              & 77.47          & 56.40          & 87.50          & 55.38          \\
                              & Mix           & 41.35           & 51.82               & 19.30              & 77.19          & 55.80          & 86.90          & 55.39          \\
                              & Family-Strong       & 42.44           & 53.37               & 20.40              & 77.94          & 57.10          & 88.30          & 56.59          \\
                              & CAR               & 43.03           & 53.81               & 20.90              & 78.42          & \textbf{58.10} & 88.80          & 57.17          \\
                              & PerSyn (Ours)      & \textbf{44.16}  & \textbf{55.14}      & \textbf{22.30}     & \textbf{79.09} & 57.80          & \textbf{90.10} & \textbf{58.09} \\ \addlinespace[2pt] \hdashline[1pt/1pt] \addlinespace[2pt]
\multirow{5}{*}{Llama-3.2-3B} & Strong           & 27.89           & 42.31               & 10.80              & 34.59          & 21.40          & 52.20          & 31.37          \\
                              & Mix           & 29.78           & 43.56               & 12.00              & 34.17          & 20.50          & 51.50          & 31.75          \\
                              & Family-Strong       & 28.25           & 42.63               & 11.10              & 33.83          & 19.50          & 50.80          & 30.85          \\
                              & CAR               & 30.53           & 44.32               & 11.80              & 35.91          & 22.80          & 53.60          & 32.99          \\
                              & PerSyn (Ours)      & \textbf{32.31}  & \textbf{46.15}      & \textbf{12.60}     & \textbf{38.15} & \textbf{24.50} & \textbf{55.30} & \textbf{34.81} \\\bottomrule[1pt]\bottomrule[1.5pt]
\end{tabular}
\vspace{-0.2cm}
\caption{Results of baseline methods and our \textit{PerSyn} strategy evaluated on six benchmarks with five student models from different families. See \S\ref{para:baselines} for details about the \strong, \mix, \family, and \car baselines.}
\vspace{-0.2cm}
\label{tab:main_result}
\end{table*}

\paragraph{Student models.}
In our main experiments, we employ five student models from Qwen2.5~\citep{Qwen2.5}, Gemma-2~\citep{gemma2}, and Llama-3.2~\citep{llama32} model families, which are Qwen2.5-0.5B, Qwen2.5-1.5B, Gemma-2-2B, Qwen2.5-3B, and Llama-3.2-3B.

\paragraph{Teacher models.}
We consider teacher models of various sizes and families for distillation in two settings.
Specifically, in the instruction tuning, we employ 19 teacher models from Qwen2~\citep{qwen2}, Qwen2.5~\citep{Qwen2.5}, Llama-3/3.1~\citep{llama3}, Gemma-2~\citep{gemma2}, and Phi-3~\citep{Phi3} model families.
For math reasoning, we employ 15 teacher models from Mistral~\citep{Mistral7}, Gemma-2~\citep{gemma2}, Llama-3.1/3.3~\citep{llama3}, Qwen2.5~\citep{Qwen2.5}, Qwen2.5-Math~\citep{qwen_math}, Qwen3~\citep{qwen3}, and DeepSeek-R1~\citep{deepseek-r1} model families.
An overview of the teacher models for the two setting is presented in Table~\ref{tab:teacher_models} of Appendix~\ref{appendix:teacher_models}.

\paragraph{Baselines.}
\label{para:baselines}
We compare our proposed \textit{PerSyn} against several baseline methods, all of which are listed below.
\textbf{1) Strong}: a straightforward approach that uses a single strongest LLM as the teacher to synthesize data; 
\textbf{2) Mix}: mix distillation~\citep{learnability_gap} that derives the synthetic dataset by mixing responses from weak and strong teacher models;
\textbf{3) Family-Strong}: a potential baseline based on the finding of~\citep{stronger_are_not}, which suggests that learning from strong teacher model within the same family as student model can improve distillation effectiveness;
\textbf{4) CAR}: a metric proposed by~\citet{stronger_are_not} that selects a single teacher model which strikes a dedicate balance between response quality and compatibility.
The specific teacher models used by the baselines for each student model in different tasks are shown in Table~\ref{tab:baselines_instruction_tuning} and Table~\ref{tab:baselines_math_reasoning} of appendix~\ref{appendix:implementation_details}.

\paragraph{Implementation.}
We train the student models using the LLaMA-Factory~\citep{llamafactory} framework. 
Student models up to 14B parameters are trained with full-parameter fine-tuning, while those larger than 14B are fine-tuned with LoRA.
For instruction tuning, we employ the state-of-the-art reward model Skywork-Reward-Llama-3.1-8B from RewardBench~\citep{lambert2025rewardbench} to obtain the quality rewards.\footnote{In the math reasoning setting, the quality reward is binary: 1 for correct answers and 0 for incorrect ones.}
We also conduct additional experiments to study the impact of weak versus strong reward models.
See Appendix~\ref{appendix:implementation_details} for details about training setup and reward models experiments.

\subsection{Performance of \textit{PerSyn}}
\label{sec:main_result}
Table~\ref{tab:main_result} reports the results of baselines and our proposed \textit{PerSyn} strategy across different benchmarks and student models. 
The results demonstrate that \textit{PerSyn} consistently outperforms all baselines in both instruction tuning and math reasoning settings. 
For instance, on Qwen2.5-3B, \textit{PerSyn} surpasses the \strong baseline by 2.9\%, 7.6\%, and 8.7\% on SVAMP, TruthfulQA, and IFEval, respectively, indicating that the strongest model is not always the best teacher for small student models. 
Relative to the strong baseline \car on Llama-3.2-3B, \textit{PerSyn} achieves gains of 4.1\% on TruthfulQA, 5.8\% on IFEval, and a more substantial 7.5\% on MATH. 
Based on these observations, we conclude that: 

\begin{tcolorboxinsight}
    \textit{PerSyn}, by jointly considering both learnability and quality to find the optimal teacher for each prompt tailored to the student model, leads to more effective student learning.
    \label{insight:1}
\end{tcolorboxinsight}

\subsection{Further Analysis}
\label{sec:further_analysis}
\paragraph{Ablation Study of 
\label{para:ablation_study}
\textit{PerSyn}}

\begin{figure}[!h]
    \centering \centerline{\includegraphics[width=0.9\columnwidth]{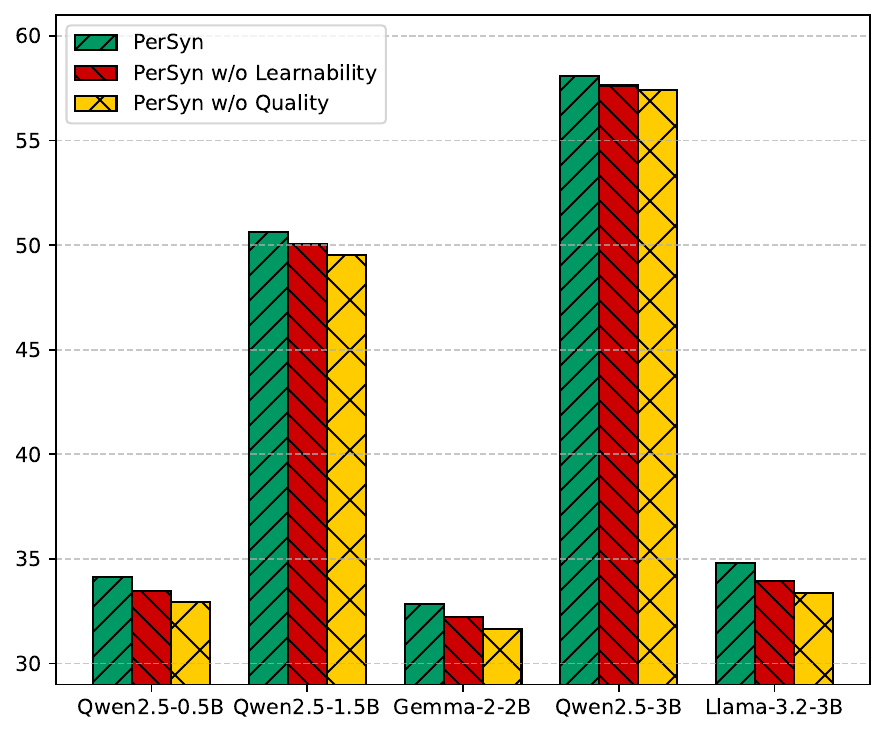}}
    \vspace{-0.1cm}
    \caption{Average results of the ablation studies on \textit{PerSyn} across all benchmarks. ``w/o'' denotes the exclusion of a specific reward term from \textit{PerSyn} when assigning prompts to teachers.}
    \label{fig:ablation_study}
\end{figure}

We conduct ablation studies to examine the roles of learnability and quality within \textit{PerSyn}. 
Specifically, we evaluate the performance of \textit{PerSyn} when either learnability or quality is excluded.
As shown in Fig.~\ref{fig:ablation_study}, the results demonstrate consistent performance drops for both ``\textit{PerSyn} w/o Learnability'' and ``\textit{PerSyn} w/o Quality'' across all student models. 
Moreover, the performance degradation is more pronounced when quality is excluded, suggesting that relying solely on high-learnability data, i.e., knowledge that is easy for the student to absorb, offers limited improvement. 
These observations indicate that:
\begin{tcolorboxinsight}
    Jointly considering both learnability and quality yields better performance than considering either alone, with quality playing a more critical role than learnability in \textit{PerSyn}.
    \label{insight:2}
\end{tcolorboxinsight}

\paragraph{\textit{PerSyn} on Larger Model Scales}
\begin{figure}[!th]
    \centering \centerline{\includegraphics[width=0.9\columnwidth]{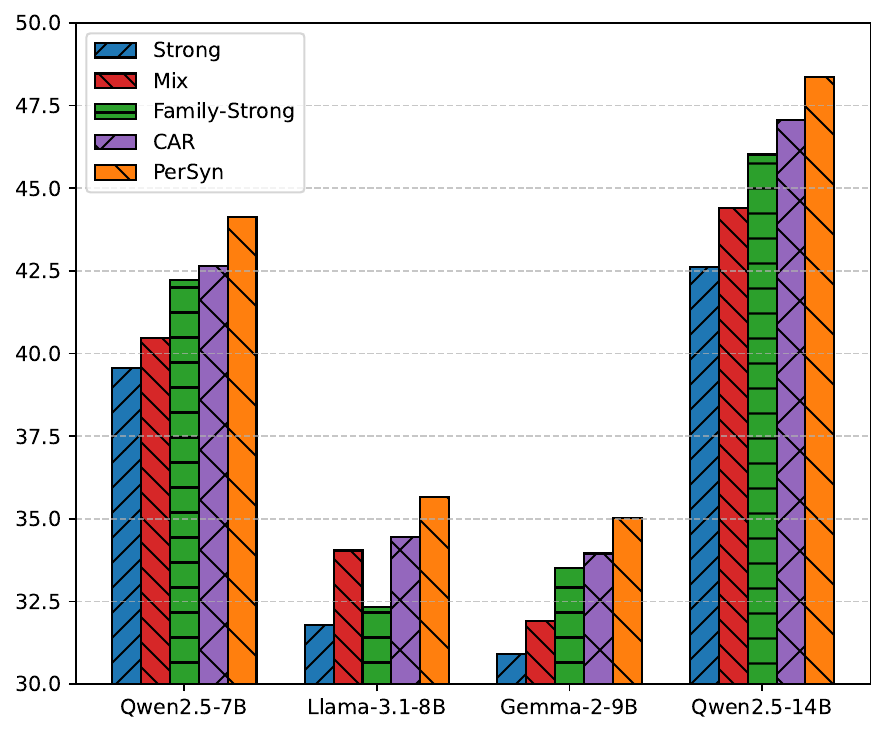}}
    \caption{The average results of baselines and \textit{PerSyn} on four larger-scale student models spanning three model families in the instruction tuning setting. Detailed results are provided in Appendix~\ref{appendix:larger_size_result}.}
    \vspace{-0.2cm}
   \label{fig:larger_scale_results}
\end{figure}

Having verified the effectiveness of \textit{PerSyn} on small student models, we further evaluate its generalization to larger-scale models in the instruction tuning setting. 
As shown in Fig.~\ref{fig:larger_scale_results}, \textit{PerSyn} consistently outperforms all baselines in terms of average performance. 
In particular, compared to the strong baseline \car, \textit{PerSyn} achieves average improvements of 3.4\%, 3.6\%, 3.1\%, and 2.7\% on Qwen2.5-7B, Llama-3.1-8B, Gemma-2-9B, and Qwen2.5-14B, respectively. 
These findings demonstrate the broad effectiveness of \textit{PerSyn} across diverse model scales and families.

\begin{figure}[!h]
    \centering 
    \centerline{\includegraphics[width=0.95\columnwidth]{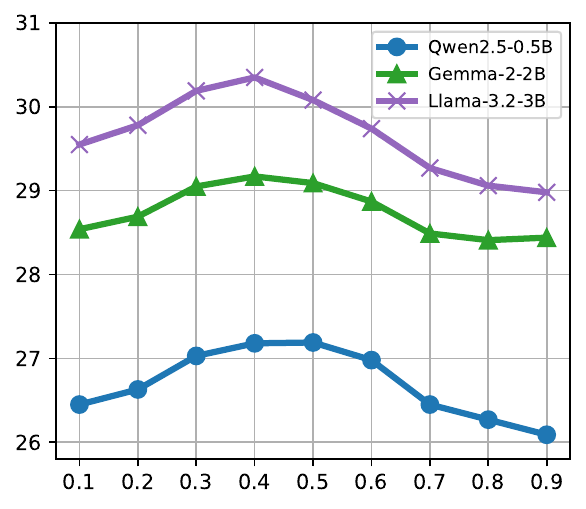}}
    \vspace{-0.2cm}
    \caption{The average results across different $\alpha$ values for three student models from distinct model families in the instruction tuning setting.}
    \label{fig:different_ratio}
    \vspace{-0.2cm}
\end{figure}

\paragraph{Suitable $\alpha$ for \textit{PerSyn}}
To determine the suitable value for $\alpha$ in Eq.~\ref{eq:overall_score} within \textit{PerSyn}, we conduct experiments with $\alpha$ ranging from 0.1 to 0.9. 
The results in Fig.~\ref{fig:different_ratio} exhibit a rising trend initially, reaching a peak at $\alpha = 0.4$, and then gradually declining, suggesting that quality is more important than learnability, consistent with the Finding~\ref{insight:2} we derive in \S\ref{para:ablation_study}. 
A similar trend is observed across three student models from different families. 
Therefore, we set $\alpha = 0.4$ by default in all experiments.

\paragraph{Performance of \textit{PerSyn} Router}
\label{par:router_performance}
\begin{figure}[!t]
    \centering \centerline{\includegraphics[width=0.85\columnwidth,height=5.5cm]{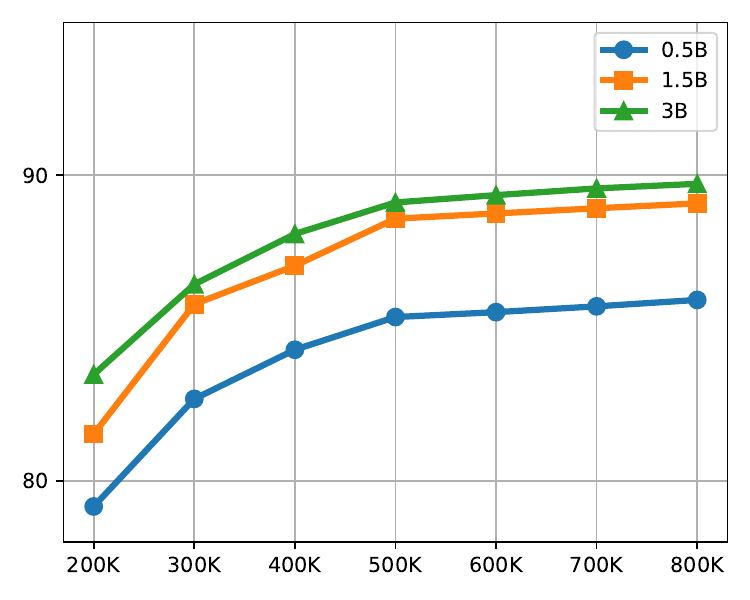}}
    \caption{The performance of the \textit{PerSyn} router for the Qwen2.5-3B student model across different backbone model sizes and pairwise training dataset sizes in the instruction tuning setting. Notably, 500K pairwise training samples, which can be constructed from only 2.5K parallel teacher responses, are sufficient to obtain an effective \textit{PerSyn} router. Similar observations in math reasoning setting are provided in Appendix~\ref{appendix:details_of_router}.}
\label{fig:qwen2.5_instruct_router_acc}
\end{figure}

We conduct additional experiments to study the impact of pairwise training dataset size and backbone model size on the performance of the \textit{PerSyn} router. 
In these experiments, Qwen2.5 serves as the backbone model for \textit{PerSyn} router. 
Fig.~\ref{fig:qwen2.5_instruct_router_acc} presents the Hit@3 performance\footnote{Hit@3 denotes the proportion of cases where the teacher model assigned by the \textit{PerSyn} router falls within the top-3 ground-truth teachers.} of the router across different backbone model sizes (0.5B to 3B) and pairwise training dataset sizes (200K to 800K) in the instruction tuning setting. 
The results show that router performance initially improves with increasing training data and then stabilizes once the dataset exceeds 500K (constructed by 2.5K prompts with parallel teacher responses, see Appendix~\ref{appendix:details_of_router} for more details). 
Similar trends are observed across all three model sizes. 
We also note that the performance of the 1.5B model is comparable to that of the 3B model. 
Based on these observations, unless otherwise specified, the \textit{PerSyn} router is obtained by training Qwen2.5-1.5B on 2.5K prompts with parallel teacher responses by default.

To further validate the effectiveness of the \textit{PerSyn} router, we conduct an additional experiment comparing it with the \textit{Oracle} router, which directly leverages ground-truth reward (as defined in Eq.~\ref{eq:overall_score}) to route each prompt to its optimal teacher model. 
Table~\ref{tab:oracle_router} reports the average results in the instruction tuning setting across different student models. 
The results show that the \textit{PerSyn} router achieves performance comparable to, or even exceeding, that of the \textit{Oracle} router (similar observations in the math reasoning setting are provided in Table~\ref{tab:oracle_router_math}). 
Notably, the \textit{Oracle} router requires parallel teacher responses for the entire prompt set. 
In contrast, the \textit{PerSyn} router only requires 2.5K parallel teacher responses, making it significantly more efficient.

\begin{table}[!t]
\centering
\setlength\tabcolsep{5pt}
\fontsize{10}{11}\selectfont
\begin{tabular}{lcc}
\toprule[1.5pt]
\toprule[1pt]
             & \multicolumn{1}{l}{\textit{PerSyn} Router} & \multicolumn{1}{l}{\textit{Oracle} Router} \\ \addlinespace[2pt] \hline \addlinespace[2pt]
Qwen2.5-0.5B & 27.18                             & 27.63                             \\
Qwen2.5-1.5B & 33.72                             & 34.36                             \\
Gemma-2-2B   & 29.17                             & 28.84                             \\
Qwen2.5-3B  & 40.53                             & 41.02                             \\
Llama-3.2-3B & 30.35                             & 30.18                             \\ \bottomrule[1pt]\bottomrule[1.5pt]
\end{tabular}
\caption{The average performance of different student models in the instruction tuning setting using the \textit{PerSyn} router and the \textit{Oracle} router.}
\label{tab:oracle_router}
\end{table}

\paragraph{Teacher Models Allocated by \textit{PerSyn}}

\begin{figure}[!t]
    \centering \centerline{\includegraphics[width=\columnwidth]{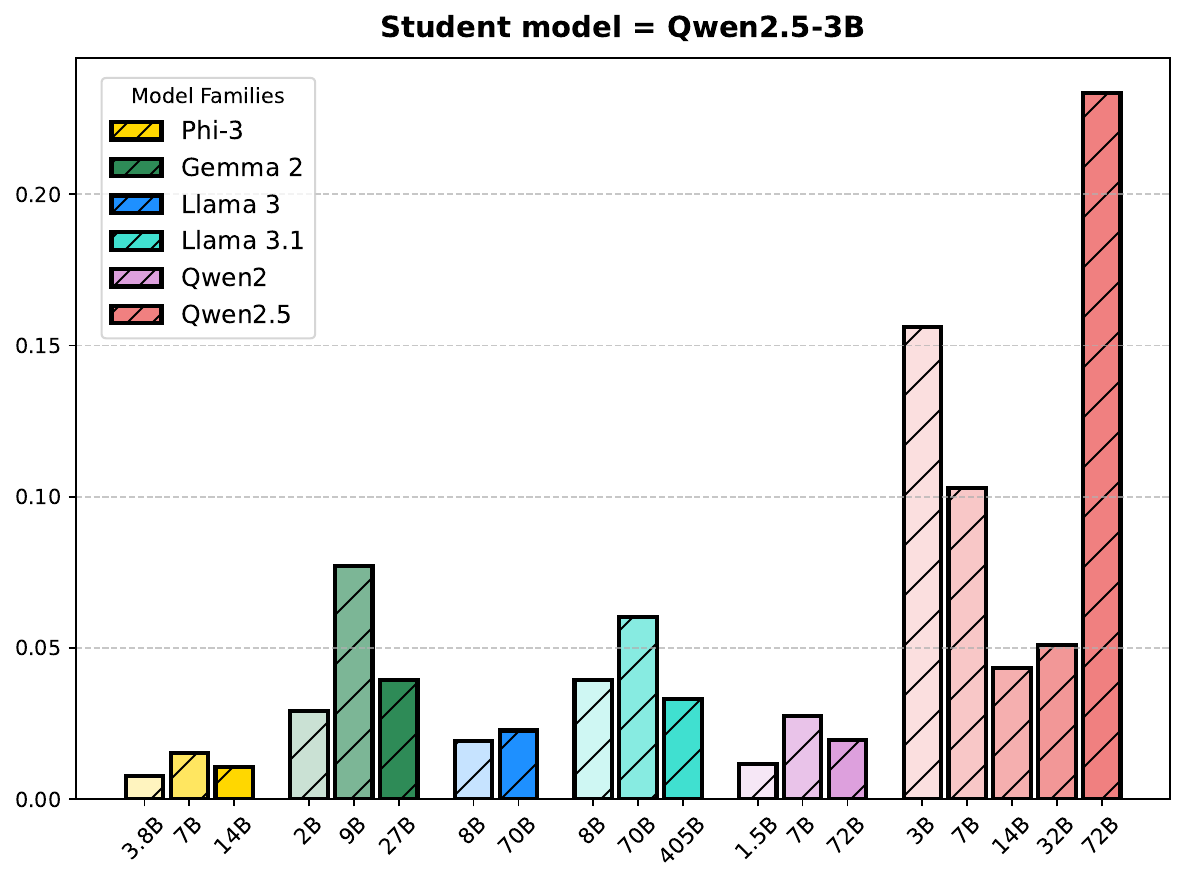}}
    \caption{Prompt allocation ratios assigned by \textit{PerSyn} across different teacher models for Qwen2.5-3B student in the instruction tuning setting. Colors indicate different model families, and darker shades correspond to larger teacher models within the same family.}
    \label{fig:qwen2.5_instruct_teacher_distribution}
\end{figure}

\begin{figure}[!t]
    \centering \centerline{\includegraphics[width=\columnwidth]{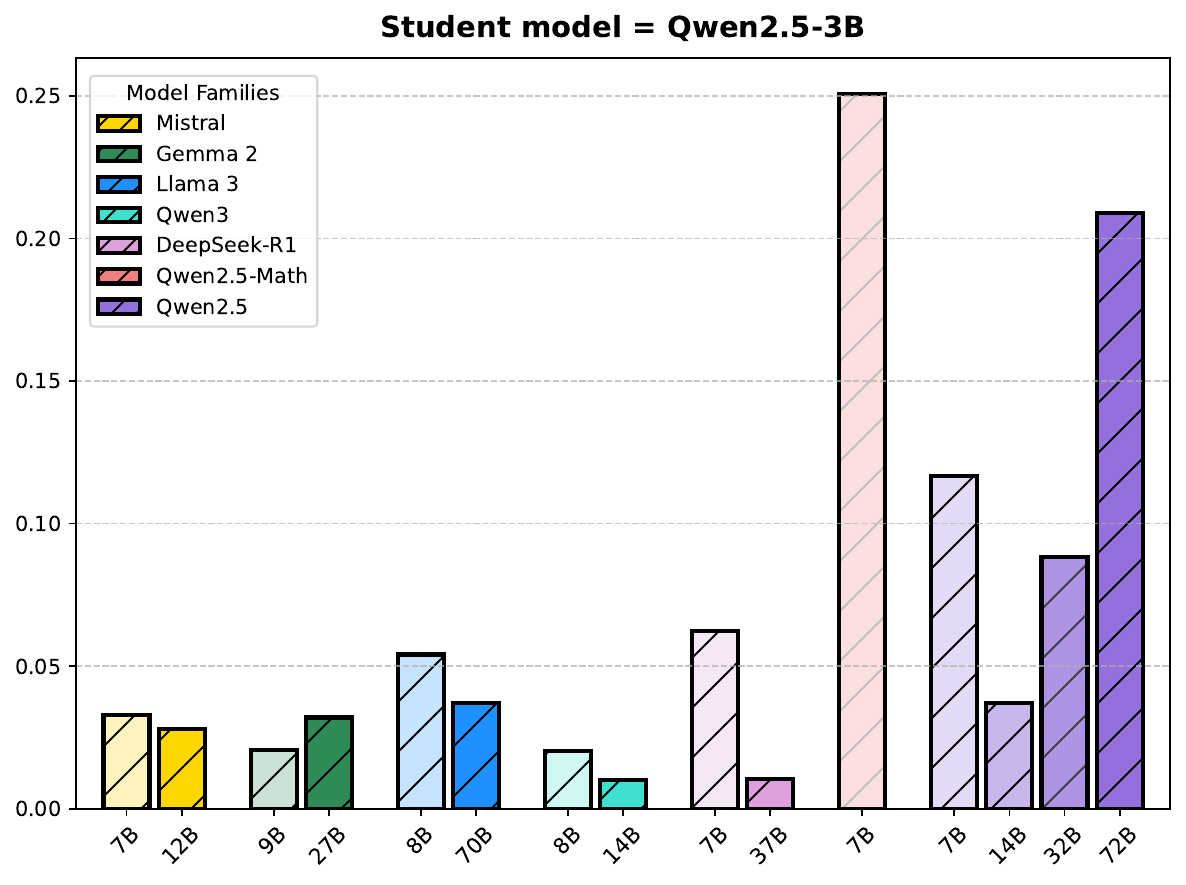}}
    \vspace{-0.2cm}
    \caption{Prompt allocation ratios assigned by \textit{PerSyn} across different teacher models for Qwen2.5-3B student in the math reasoning setting. Colors indicate different model families, and darker shades correspond to larger teacher models within the same family. Note that DeepSeek-R1 and Qwen3 are Long-CoT models, while the remaining are Short-CoT models.
}\label{fig:qwen2.5_math_teacher_distribution}
\vspace{-0.2cm}
\end{figure}

To delve deeper, we visualize the prompt allocation ratios assigned by \textit{PerSyn} router across different teacher models for Qwen2.5-3B under instruction tuning and math reasoning settings in Fig.~\ref{fig:qwen2.5_instruct_teacher_distribution} and Fig.~\ref{fig:qwen2.5_math_teacher_distribution}. 
As shown in Fig.~\ref{fig:qwen2.5_instruct_teacher_distribution}, smaller teacher models, such as Qwen2.5-3B-Instruct, receive higher allocation compared to larger models, including Qwen2.5-7B/14B/32B-Instruct and even Llama-3.1-405B-Instruct. 
A similar trend is observed in math reasoning (Fig.~\ref{fig:qwen2.5_math_teacher_distribution}), where Qwen2.5-7B-Instruct has higher allocation than Qwen2.5-14B/32B-Instruct.
See Appendix~\ref{appendix:teacher_model_distribution} for the prompt allocation ratios of other student models. 
Based on these observations, we derive the following conclusion: 
\begin{tcolorboxinsight}
    Larger teacher models, despite their superior performance, are not always the optimal teacher for small student models; small teachers are often more suitable.\label{insight:3}
\end{tcolorboxinsight}

Interestingly, Qwen2.5-72B-Instruct consistently receives high allocation across student models in both settings, suggesting it is a strong and versatile teacher.

In addition, Fig.~\ref{fig:qwen2.5_math_teacher_distribution} shows that teacher models producing Long-CoT responses account for only a small portion of the allocated prompts compared to Short-CoT models\footnote{In this paper, we refer to models that generate Long-CoT or Short-CoT responses as Long-CoT and Short-CoT models, respectively.}. Similar trends for other student models are reported in Appendix~\ref{appendix:teacher_model_distribution}. 
To examine their necessity of Long-CoT models, we conduct an experiment where prompts originally assigned to Long-CoT models are forcibly reassigned to a strong Short-CoT teacher (Qwen2.5-Math-7B-Instruct), while keeping the allocation of all other prompts unchanged.
We then construct the synthetic dataset based on this modified allocation and use it to train the Qwen2.5-3B student model.
The result shows that this replacement led to a 1.3\% drop in average performance compared to the original \textit{PerSyn} allocation.
Moreover, Qwen2.5-Math-7B-Instruct was able to correctly answer only 7.4\% of the prompts that were initially assigned to Long-CoT models.
These findings indicate that:
\begin{tcolorboxinsight}
    While Long-CoT models are not optimal teachers for small student models in most cases, they remain necessary for handling certain complex prompts.
    \label{insight:4}
\end{tcolorboxinsight}

Finally, Table~\ref{tab:main_result} shows that training student models on datasets consisting entirely of Long-CoT responses (\strong baseline) results in degraded performance relative to \textit{PerSyn}, indicating that excessive reliance on Long-CoT data is detrimental.
Further analysis of generated outputs shows that models trained with \strong often produce repetitive reasoning without termination, leading to incorrect answers. 
In contrast, models trained with \textit{PerSyn} can generate suitably extended reasoning paths and produce correct results. 
This aligns with the observations reported in~\citet{learnability_gap}.

\section{Related Work}
\label{sec:related}
\paragraph{Large Language Models for Data Synthesis}
Due to the high cost of human data annotation, recent research has turned to Large Language Models (LLMs) for synthetic data generation as a practical alternative~\citep{long2024llms,tan2024large}.
This line of work has demonstrated promising results across diverse domains.
For example, LLMs can generate instruction-response pairs in instruction tuning~\citep{alpaca,wang2024bpo,hescaling,yuan2025improving} and elaborate rationales for reasoning tasks~\citep{shridhar-etal-2023-distilling,li-etal-2025-learning-committee, deepseek-r1}. Other task-specific data synthesis include dialogue generation~\citep{sun2024fostering,wang2025toolflow,tan2025prospect}, code generation~\citep{codellama, majumdar-etal-2025-genetic}, reinforcement learning~\citep{liang2025sws,liang2025beyond}, data augmentation~\citep{wang2024surveydatasyn,liang2024task}, and retrieval-augmented generation~\cite{xu-etal-2025-simrag,li2025smoa,yang2025quantifying}.
In this work, we mainly conduct experiments on two prevalent data synthesis scenarios: instruction tuning and math reasoning.

\paragraph{Distillation with Synthetic Data}
To improve efficiency, previous studies aimed to use the synthetic data generated by large and powerful LLMs to train smaller models~\citep{vicuna,busbridgedistillation}.
However, recent works have demonstrated that stronger models are not always stronger teacher models for data generation~\citep{kim2024evaluating,stronger_are_not,learnability_gap,chen2025unveiling}. Specifically, ~\citet{learnability_gap} showed that small student models struggle to learn from strong reasoners due to the learnability gap and introduced a mixed distillation strategy to mitigate this gap.
~\citet{stronger_are_not} found that learning from response generators within the same model family yields higher performance and propose a metric to select the best teacher by measuring the quality and compatibility. 
However, these methods are inefficient and ineffective, as they require generating parallel teacher responses and cannot produce a synthetic dataset that is optimal at the sample level.
In this work, we aim to assign each prompt to the most suitable teacher model for data synthesis, thereby constructing the final personalized dataset.

\section{Conclusion}
\label{sec:conclusion}
In this work, we propose \textit{PerSyn}, a new strategy that constructs personalized synthetic data for a specific student model to help it learn more effectively.
Unlike previous work that employ a single selected teacher model to synthesize data for the entire prompt set, \textit{PerSyn} operates in a more fine-grained manner: it assigns each prompt to its optimal teacher model for synthesis, based on both the student model's learnability and the teacher model's response quality.
Furthermore, \textit{PerSyn} transfers the synthesis paradigm from the conventional ``Generate then Select'' to ``Route then Generate'' by introducing a router-guided mechanism.
Extensive experiments demonstrate that synthetic data constructed by \textit{PerSyn} effectively facilitates the learning of student models, achieving state-of-the-art performance across various scales and model families in both instruction tuning and math reasoning scenarios. 
Our comprehensive analysis also offers valuable insights for future research.

\section*{Limitations}
\label{sec:limitation}
While \textit{PerSyn} demonstrates strong effectiveness in both instruction tuning and math reasoning, it remains unclear whether \textit{PerSyn} can generalize to other scenarios, such as code generation, multi-modal understanding, and other specialized domains.
Furthermore, our experiments are limited to student models with up to 14B parameters, and we have not evaluated larger LLMs (e.g., 32B or 70B) due to computational constraints. We leave these for future work.

\section*{Ethical Considerations}
\label{sec:ethics}
All datasets used in our work are publicly released under open licenses, and all models employed in distillation or generation are open-source and licensed for research use. We strictly follow the licensing terms and usage policies of these resources.

\section*{Acknowledgement}
This work was supported in part by the Theme-based Research Scheme (TRS) project T45-701/22-R of the Research Grants Council of Hong Kong, and in part by the AVNET-HKU Emerging Microelectronics and Ubiquitous Systems (EMUS) Lab.

\bibliography{custom}

\newpage
\onecolumn
\appendix
\section{Appendix}
\label{sec:appendix}
\subsection{Datasets and Evaluation}
\label{appendix:dataset_details}
\begin{table*}[!htb]
\centering
\setlength\tabcolsep{4pt}
\fontsize{10}{12}\selectfont
\begin{tabular}{lccc}
\toprule[1.5pt]
\toprule[1pt]
\textbf{Dataset} & \textbf{\#Samples} & \textbf{Task Type} & \textbf{Data Type} \\
\midrule
{Magpie-100K-Generator-Zoo (Magpie-Zoo)~\citep{stronger_are_not}}       & 50K   & Instruction Tuning & Train \\
{TruthfulQA~\citep{truthfulqa}}  & 817  & Instruction Tuning     & Test \\
{LiveBench~\citep{livebench}}   & 1436  & Instruction Tuning     & Test  \\
{IFEval~\citep{ifeval}}      & 541    & Instruction Tuning & Test  \\ \addlinespace[2pt] \hdashline[1pt/1pt] \addlinespace[2pt]
{PerSyn-Math (Our)} & 10K    & Math Reasoning & Train  \\
{SVAMP~\citep{svamp}}      & 1000    & Math Reasoning & Test  \\
{MATH~\citep{math}}       & 500    & Math Reasoning & Test  \\
{GSM8K~\citep{gsm8k}}      & 1320    & Math Reasoning & Test  \\
\bottomrule[1pt]\bottomrule[1.5pt]
\end{tabular}
\caption{Overview of the datasets we used in our experiments. \#Samples indicates the number of samples per dataset. Notably, original {Magpie-Zoo} have already contained 19 parallel teacher responses for each prompt. 
We randomly sample 50K from it to construct training dataset for baselines and our \textit{PerSyn} strategy in instruction tuning setting. 
For math reasoning setting, we construct \textbf{\textit{PerSyn-Math}} dataset, which provides 10K samples with 15 parallel teacher responses.
We use \textbf{\textit{PerSyn-Math}} to construct training dataset for baselines and our \textit{PerSyn} strategy in math reasoning setting.
We use 2.5K samples with parallel teacher responses to build pairwise training data to obtain \textit{PerSyn} router in both settings (as supported by \S\ref{par:router_performance}).}
\label{tab:datasets}
\end{table*}

\paragraph{Details of Data Synthesis.} For instruction tuning, the Magpie-100K-Generator-Zoo dataset use greedy decoding to generate responses. 
For math reasoning, we set the temperature to 0.6 and use teacher models to synthesize solutions by rejection sampling. Specifically, we sample four outputs from models under 72B and two outputs from 72B or Long-CoT models. 
Each output is then verified using Math-Verify.\footnote{\url{https://github.com/huggingface/Math-Verify}} If at least one output is correct, we keep one correct solution; otherwise, we randomly keep one incorrect solution.

\paragraph{Datasets Overview.}
Table \ref{tab:datasets} presents an overview of the datasets used in our experiments. For instruction tuning setting, {TruthfulQA} is designed to measure whether a language model generate truthful answers. {LiveBench} is a monthly updated benchmark that tests LLMs on 18 diverse tasks across 6 categories.\footnote{The release option of LiveBench we used in our experiment is 2024-11-25.}
{IFEval} evaluate LLMs' capability to follow automatically verifiable natural language instructions.
For math reasoning, {SVAMP} and {GSM8K} are datasets of grade-school math word problems. And {MATH} is a more challenging benchmark that contains competition-level mathematics problems requiring complex reasoning and advanced knowledge.

\paragraph{Evaluation Setup.}
We assess the models on {LiveBench} with the official scripts\footnote{\url{https://github.com/LiveBench/LiveBench}}, reporting scores averaged across six domains: {math}, {coding}, {reasoning}, {language}, {data analysis}, and {instruction following}. {TruthfulQA} and {IFEval} are evaluated using {lm-evaluation-harness}~\citep{eval-harness}, a standard evaluation suite, and for {IFEval} we report instruction-level strict accuracy.
In the math reasoning setting, we use accuracy as the primary metric, with correctness measured by math-verify.
All benchmarks are tested in the zero-shot setting, except for {GSM8K}, which is evaluated in a 5-shot setting.
To ensure reproducibility of our empirical results, we implement greedy decoding for all benchmarks.

\newpage
\onecolumn
\subsection{Details of \textit{PerSyn} Router}
\label{appendix:details_of_router}

To determine the optimal pairwise training dataset size and model size for obtaining an effective \textit{PerSyn} router, we first reserve 1K samples as the evaluation set. 
From the remaining data, we randomly sample subsets of different sizes $N$ as training sets. 
Next, all teacher models generate parallel responses for prompts in both the training and evaluation sets. 
Then, we compute the learnability and quality rewards via Eq.~\ref{eq:overall_score} for these parallel teacher responses, which allows us to establish the ground-truth ranking for each prompt. 
The labeled training set of size $N$ is then used to construct a pairwise dataset of size $M$ for training \textit{PerSyn} routers of different model sizes, while the evaluation set is used to assess router performance.

The desired pairwise dataset size $M$ determines the required sample size $N$. 
For instance, in the instruction tuning setting, 19 teacher models\footnote{See Table~\ref{tab:teacher_models} for details of teacher models in instruction tuning and math reasoning settings.} produce 190 model combinations. 
To obtain $M=500$K pairwise samples, a subset of size $N = 500\text{K} \div 190 \approx 2.5$K prompts with parallel teacher responses suffices. 
Similarly, in the math reasoning setting, 15 teacher models yield 105 combinations; to obtain $M=250$K pairwise samples, a subset of size $N = 250\text{K} \div 105 \approx 2.5$K prompts with parallel teacher responses is sufficient. 
Table~\ref{tab:pair_wise_data_num} presents the different sample sizes $N$ and their corresponding pairwise dataset sizes $M$ for both settings.

\begin{table}[!h]
\centering
\setlength\tabcolsep{5pt}
\fontsize{9}{11}\selectfont
\begin{tabular}{lccccc}
\toprule[1.5pt]
\toprule[1pt]
 & \multicolumn{2}{c}{Instruction Tuning}  &  & \multicolumn{2}{c}{Math Reasoning}      \\ \addlinespace[2pt] \cline{2-3} \cline{5-6} \addlinespace[2pt]
 & Sample Size N & Pair-wise Size M &  & Sample Size N & Pair-wise Size M \\ \addlinespace[2pt] \hline
 & 1K           & 200K            &  & 0.5K           & 50K            \\
 & 1.5K         & 300K            &  & 1K         & 100K            \\
 & 2K           & 400K            &  & 1.5K           & 150K            \\
 & 2.5K         & 500K            &  & 2K         & 200K            \\
 & 3K           & 600K            &  & 2.5K           & 250K            \\
 & 3.5K         & 700K            &  & 3K         & 300K            \\
 & 4K           & 800K            &  & 3.5K           & 350K            \\ \bottomrule[1pt]\bottomrule[1.5pt]
\end{tabular}
\vspace{-0.1cm}
\caption{Different prompt sample sizes $N$ with parallel teacher responses and their corresponding pairwise training data sizes $M$ in the instruction tuning and math reasoning settings. To obtain $M$ pairwise training samples, each teacher model only needs to generate $N$ responses.}
\vspace{-0.3cm}
\label{tab:pair_wise_data_num}
\end{table}

\begin{figure}[!th]
    \centering \centerline{\includegraphics[width=0.4\columnwidth,height=5.3cm]{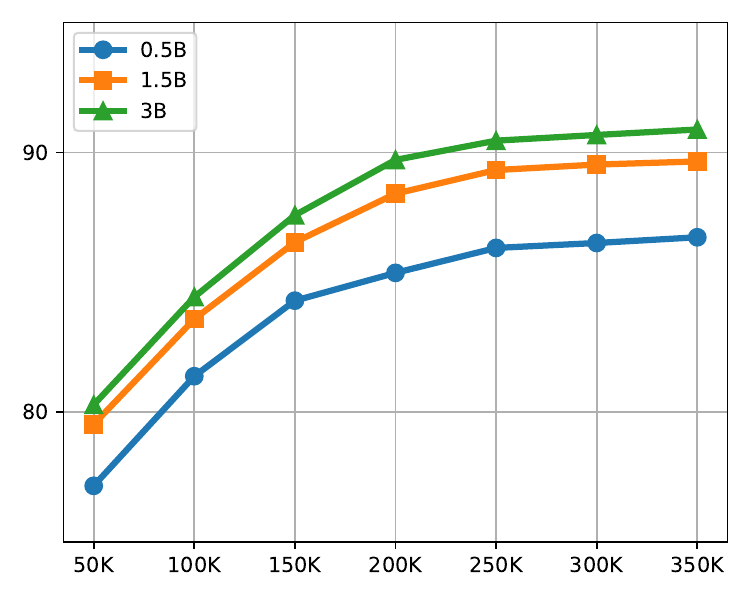}}
    \caption{The performance of the \textit{PerSyn} router for the Qwen2.5-3B student model across different backbone model sizes and pairwise training dataset sizes in the math reasoning setting. Notably, 250K pairwise training samples, which can be constructed from only 2.5K parallel teacher responses, are sufficient to obtain an effective \textit{PerSyn} router.}
    
    \label{fig:qwen2.5_math_router_acc}
\end{figure}

We present the performance of the \textit{PerSyn} router for the Qwen2.5-3B student model across different router model sizes and pairwise training dataset sizes under instruction tuning and math reasoning settings in Fig.~\ref{fig:qwen2.5_instruct_router_acc} and Fig.~\ref{fig:qwen2.5_math_router_acc}. 
In Fig.~\ref{fig:qwen2.5_instruct_router_acc}, 500K pairwise training samples, constructed from only around $N=2.5$K prompts with parallel teacher responses (see Table~\ref{tab:pair_wise_data_num}, ``Instruction Tuning'' column), are sufficient to obtain an effective \textit{PerSyn} router in the instruction tuning setting. 
Similarly, Fig.~\ref{fig:qwen2.5_math_router_acc} shows that 250K pairwise training samples, generated from approximately $N=2.5$K prompts with parallel teacher responses (see Table~\ref{tab:pair_wise_data_num}, ``Math Reasoning'' column), are sufficient to obtain an effective \textit{PerSyn} router in the math reasoning setting.

To further validate the effectiveness of the \textit{PerSyn} router, we conduct an additional experiment comparing it with the \textit{Oracle} router, which uses ground-truth rewards to directly route each prompt to its optimal teacher model. 
The average results across different student models in both instruction tuning and math reasoning settings are reported in Table~\ref{tab:oracle_router} and Table~\ref{tab:oracle_router_math}. 
In both settings, the \textit{PerSyn} router achieves performance comparable to or even surpassing that of the \textit{Oracle} router. 
Importantly, \textit{Oracle} router requires parallel teacher responses for the entire prompt set, whereas the \textit{PerSyn} router only requires 2.5K prompts with parallel responses, making it substantially more efficient.

\begin{table}[!th]
\centering
\setlength\tabcolsep{4pt}
\fontsize{10}{11}\selectfont
\begin{tabular}{lcc}
\toprule[1.5pt]
\toprule[1pt]
             & \multicolumn{1}{l}{\textit{PerSyn} Router} & \multicolumn{1}{l}{\textit{Oracle} Router} \\ \addlinespace[2pt] \hline \addlinespace[2pt]
Qwen2.5-0.5B & 41.08                             & 41.45                             \\
Qwen2.5-1.5B & 67.54                             & 68.02                             \\
Gemma-2-2B   & 36.52                             & 36.35                             \\
Qwen2.5-3B  & 75.66                             & 75.78                             \\
Llama-3.2-3B & 39.32                             & 39.13                             \\ \bottomrule[1pt]\bottomrule[1.5pt]
\end{tabular}
\caption{The average performance of different student models in the math reasoning setting using the \textit{PerSyn} router and the \textit{Oracle} router.}
\label{tab:oracle_router_math}
\end{table}

\subsection{Results of Larger-scale Student Models}
\label{appendix:larger_size_result}
\begin{table*}[!th]
\centering
\setlength\tabcolsep{4pt}
\fontsize{10}{8}\selectfont
\begin{tabular}{llccc}
\toprule[1.5pt]
\toprule[1pt]
\textbf{Student Model}        & \textbf{Strategy} & \textbf{IFEval} & \textbf{TruthfulQA} & \textbf{LiveBench} \\ \addlinespace[2pt]\hline\addlinespace[2pt]
\multirow{5}{*}{Qwen2.5-7B}   & Strong           & 52.85           & 22.80               & 43.02              \\
                              & Mix           & 53.66           & 23.50               & 44.26              \\
                              & Family-Strong       & 55.43           & 25.10               & 46.15              \\
                              & CAR               & 55.75           & 24.90               & 47.33              \\
                              & PerSyn (Ours)      & \textbf{57.12}  & \textbf{26.20}      & \textbf{49.08}     \\ \addlinespace[2pt] \hdashline[1pt/1pt] \addlinespace[2pt]
\multirow{5}{*}{Llama-3.1-8B} & Strong           & 47.71           & 14.80               & 32.87              \\
                              & Mix           & 49.82           & 16.90               & 35.42              \\
                              & Family-Strong       & 48.36           & 15.40               & 33.25              \\
                              & CAR               & 50.23           & 17.30               & 35.86              \\
                              & PerSyn (Ours)      & \textbf{51.34}  & \textbf{18.50}      & \textbf{37.13}     \\ \addlinespace[2pt] \hdashline[1pt/1pt] \addlinespace[2pt]
\multirow{5}{*}{Gemma-2-9B}   & Strong           & 45.68           & 13.60               & 33.49              \\
                              & Mix           & 46.71           & 14.40               & 34.63              \\
                              & Family-Strong       & 48.15           & 16.30               & 36.09              \\
                              & CAR               & 48.52           & 16.70               & 36.63              \\
                              & PerSyn (Ours)      & \textbf{49.66}  & \textbf{17.50}      & \textbf{37.84}     \\ \addlinespace[2pt] \hdashline[1pt/1pt] \addlinespace[2pt]
\multirow{5}{*}{Qwen2.5-14B}  & Strong           & 55.84           & 25.70               & 46.34              \\
                              & Mix           & 57.26           & 27.10               & 48.85              \\
                              & Family-Strong       & 59.07           & 28.40               & 50.59              \\
                              & CAR               & 60.32           & 28.80               & 52.06              \\
                              & PerSyn (Ours)      & \textbf{62.15}  & \textbf{29.50}      & \textbf{53.47}     \\\bottomrule[1pt]\bottomrule[1.5pt]
\end{tabular}
\caption{Results of baseline methods and our \textit{PerSyn} strategy evaluated on four larger-scale student models from different families. See \S~\ref{para:baselines} for details about the \strong, \mix, \family, and \car baselines.}
\label{tab:larger_size_main_result}
\end{table*}

\newpage
\onecolumn
\subsection{Prompt Allocation Ratios Assigned by \textit{PerSyn} across Different Teacher Models}
\label{appendix:teacher_model_distribution}
\begin{figure}[!h]
\begin{minipage}{0.5\textwidth} 
    \includegraphics[width=\textwidth]{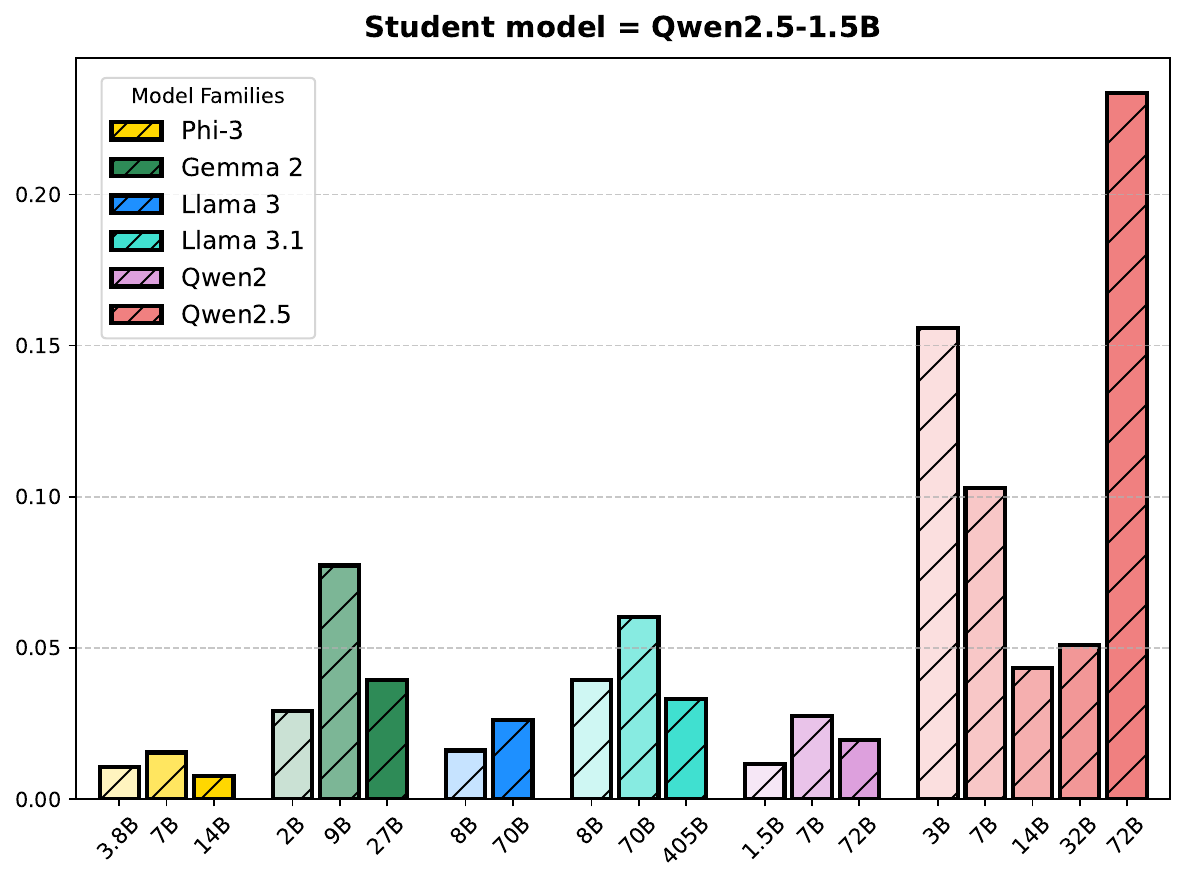} 
    \vspace{10pt} 
    
    \includegraphics[width=\textwidth]{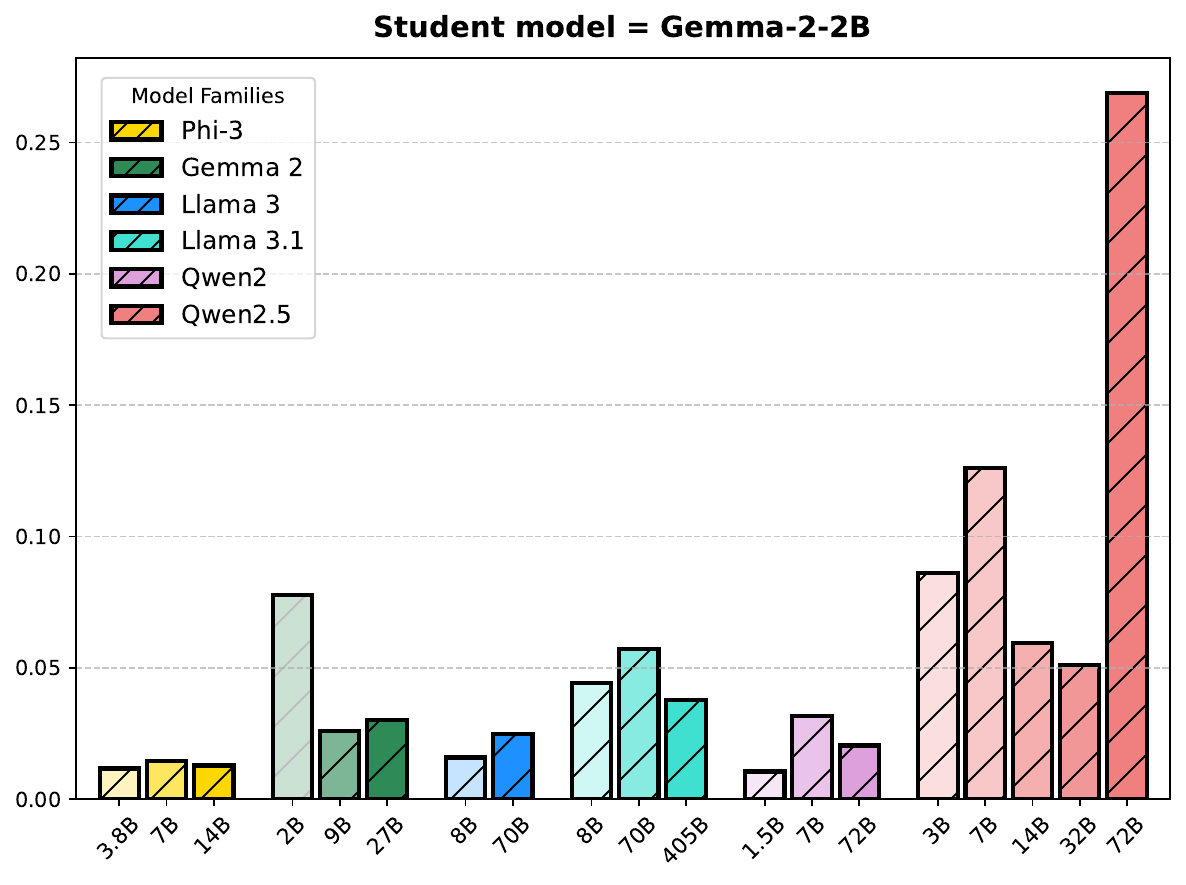}
    \vspace{10pt}
    
    \includegraphics[width=\textwidth]{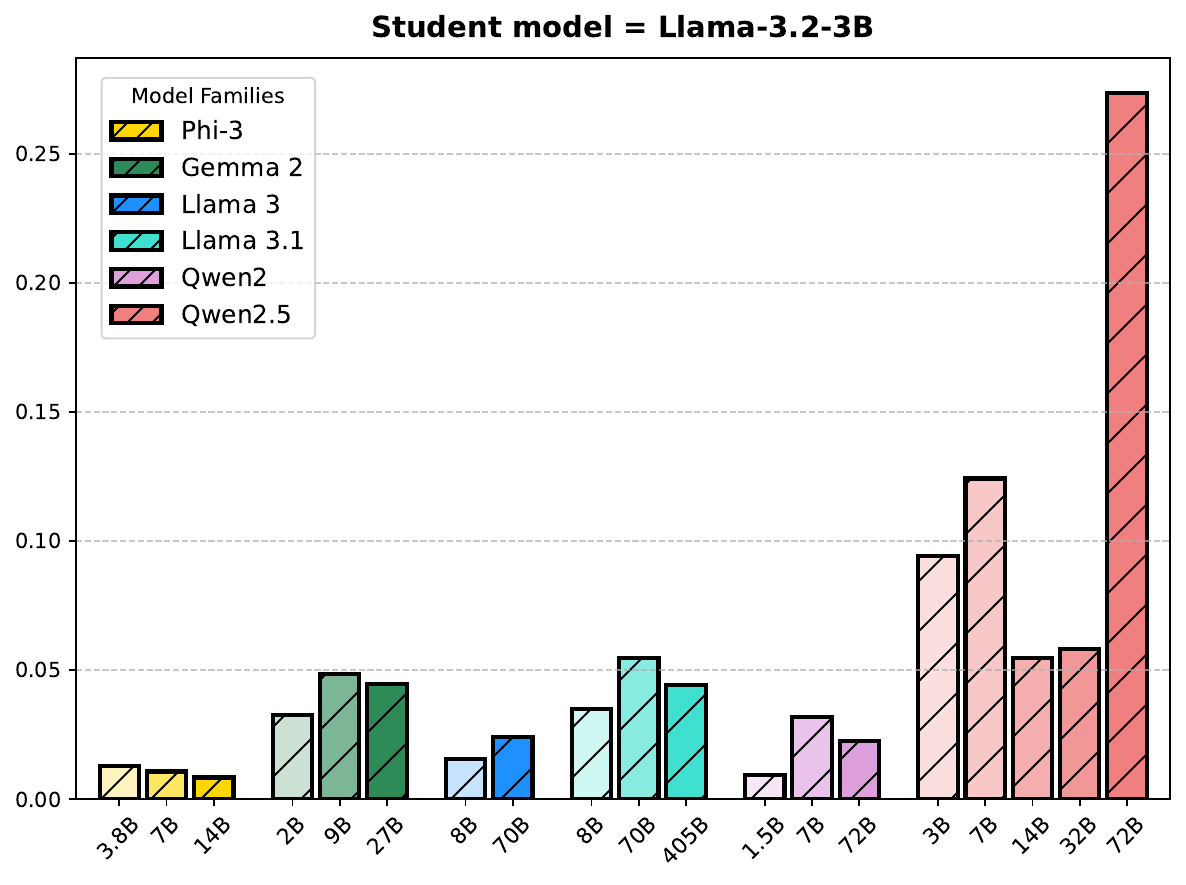}
\end{minipage}
\begin{minipage}{0.5\textwidth} 
    \includegraphics[width=\textwidth]{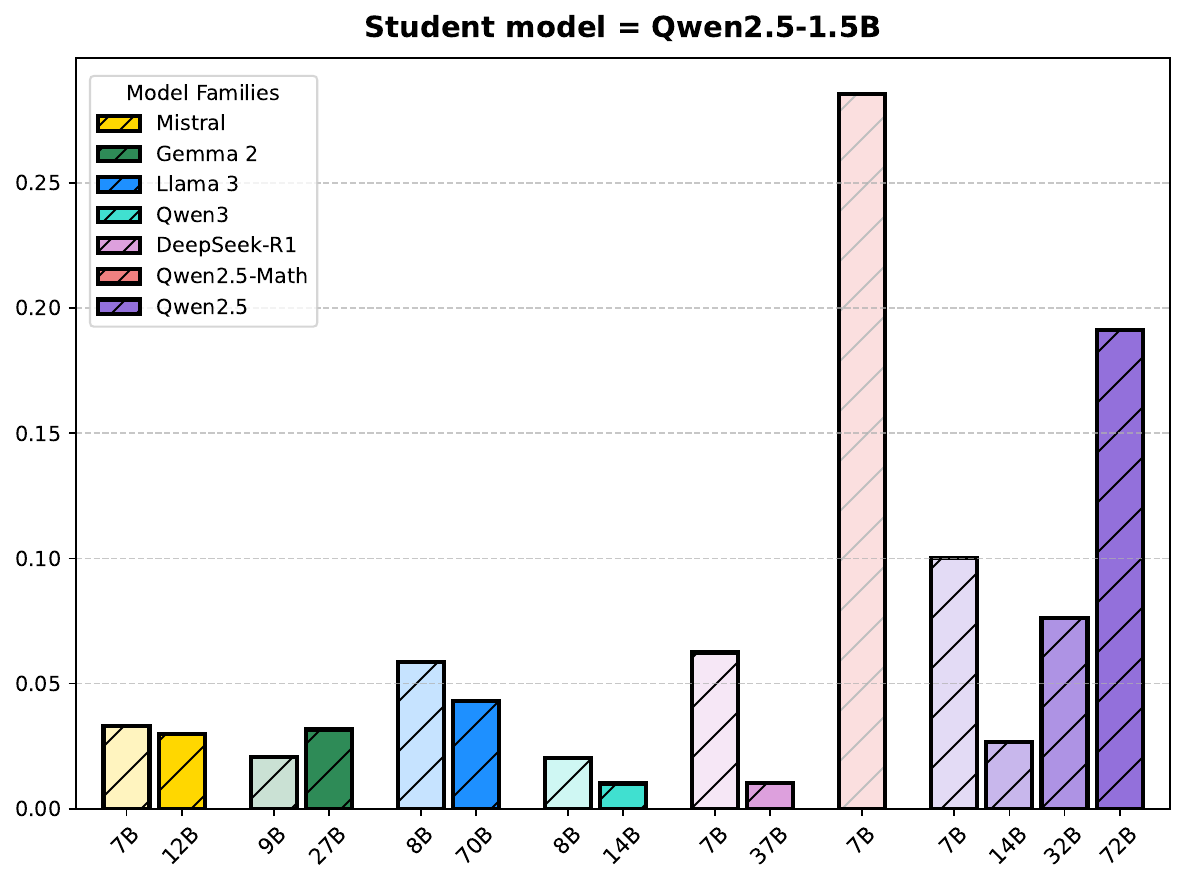} 
    \vspace{10pt}
    
    \includegraphics[width=\textwidth]{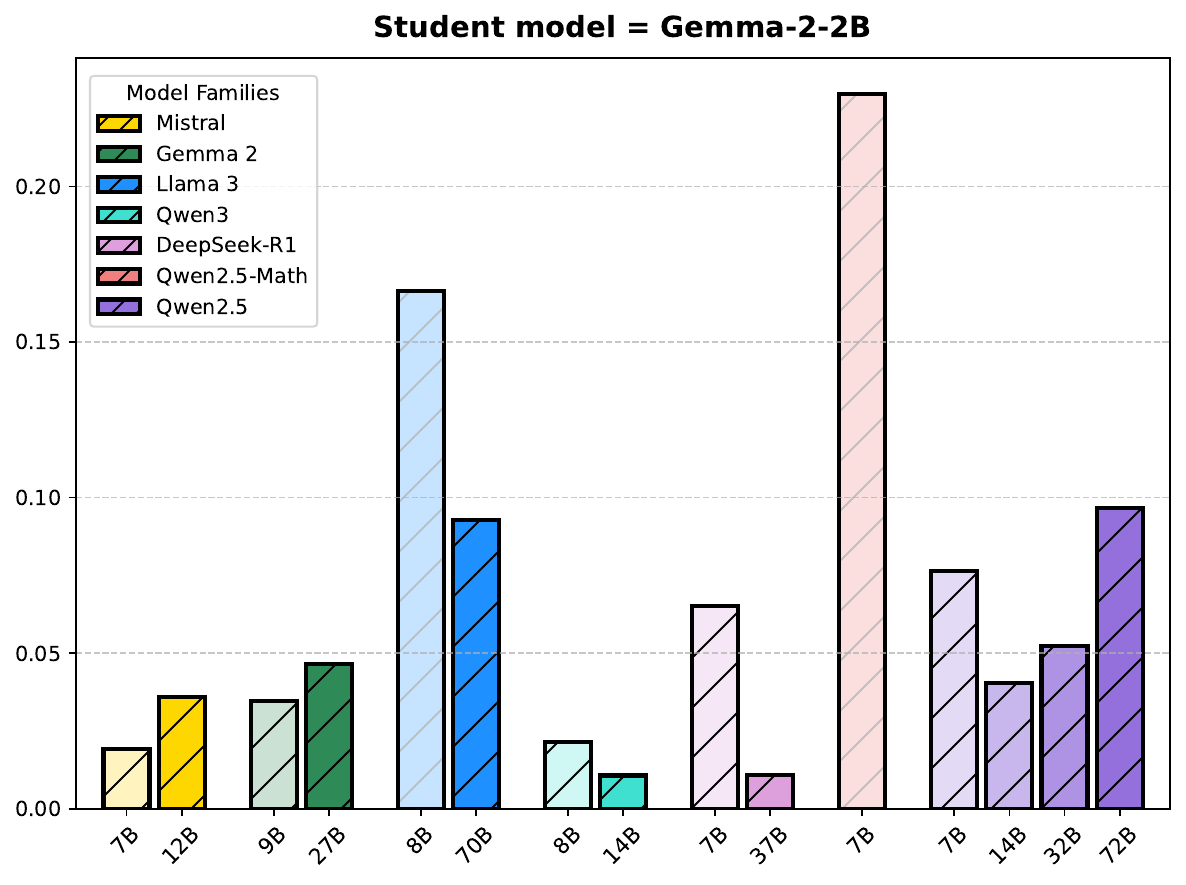}
    \vspace{10pt}
    
    \includegraphics[width=\textwidth]{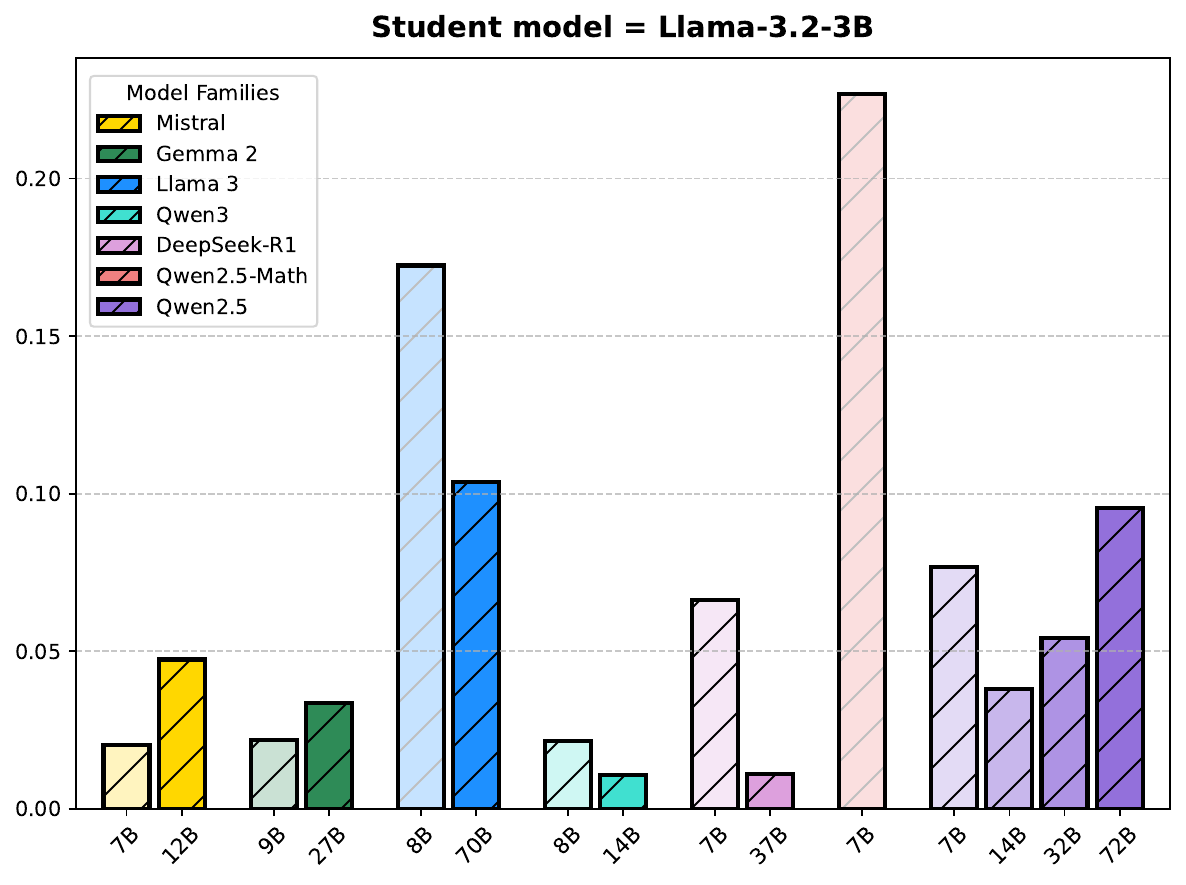}
\end{minipage}
\caption{Prompt allocation ratios assigned by \textit{PerSyn} across different teacher models for Qwen2.5-1.5B, Gemma-2-2B, and Llama-3.2-3B student models in the instruction tuning (left) and math reasoning (right) settings. Colors indicate different model families, and darker shades correspond to larger teacher models within the same family. Note that DeepSeek-R1 and Qwen3 are Long-CoT models, while the remaining are Short-CoT models.}
\end{figure}

\newpage
\onecolumn
\subsection{Teacher Models of Two Settings}
\label{appendix:teacher_models}

Table~\ref{tab:teacher_models} presents an overview of the teacher models used in our two settings. The teacher model pool for instruction tuning includes 19 models from 6 different families, with sizes ranging from 1.5B to 405B.
For math reasoning, we use 15 teacher models from 7 different families. 
Notably, beyond conventional teacher model choices, we also include models specialized for math reasoning, covering RL-trained reasoning models ({Qwen3} and {DeepSeek-R1}), a backbone model pretrained on math data ({Qwen2.5 Math}), and a model distilled with Long-CoT rationales ({DeepSeek-R1-Distill-Qwen-7B}).
It should be noted that the 37B DeepSeek-R1 model denotes its number of activated parameters, whereas the full model contains 685B parameters in total.

\begin{table*}[htbp]
\renewcommand{\arraystretch}{1.2}
\setlength\tabcolsep{5pt}
\fontsize{10}{11}\selectfont
\begin{minipage}[t]{0.4\textwidth}
\centering 
\begin{tabular}{@{} l l c @{}}
\toprule[1.5pt]
\toprule[1pt]
\textbf{Model Family} & \textbf{Model ID} & \textbf{Size} \\
\addlinespace[2pt] \hdashline[1pt/1pt] \addlinespace[2pt]
\multirow{3}{*}{\shortstack[l]{\textbf{Qwen2}}}
  & Qwen2-1.5B-Instruct & 1.5B \\
  & Qwen2-7B-Instruct   & 7B \\
  & Qwen2-72B-Instruct  & 72B \\
\addlinespace[2pt] \hdashline[1pt/1pt] \addlinespace[2pt]
\multirow{5}{*}{\shortstack[l]{\textbf{Qwen2.5}}}
  & Qwen2.5-3B-Instruct   & 3B \\
  & Qwen2.5-7B-Instruct   & 7B \\
  & Qwen2.5-14B-Instruct  & 14B \\
  & Qwen2.5-32B-Instruct  & 32B \\
  & Qwen2.5-72B-Instruct  & 72B \\
\addlinespace[2pt] \hdashline[1pt/1pt] \addlinespace[2pt]
\multirow{2}{*}{\shortstack[l]{\textbf{Llama 3}}}
  & Llama-3-8B-Instruct   & 8B \\
  & Llama-3-70B-Instruct  & 70B \\
\addlinespace[2pt] \hdashline[1pt/1pt] \addlinespace[2pt]
\multirow{3}{*}{\shortstack[l]{\textbf{Llama 3.1}}}
  & Llama-3.1-8B-Instruct   & 8B \\
  & Llama-3.1-70B-Instruct  & 70B \\
  & Llama-3.1-405B-Instruct & 405B \\
\addlinespace[2pt] \hdashline[1pt/1pt] \addlinespace[2pt]
\multirow{3}{*}{\shortstack[l]{\textbf{Gemma 2}}}
  & Gemma-2-2b-it   & 2B \\
  & Gemma-2-9b-it   & 9B \\
  & Gemma-2-27b-it  & 27B \\
\addlinespace[2pt] \hdashline[1pt/1pt] \addlinespace[2pt]
\multirow{3}{*}{\shortstack[l]{\textbf{Phi-3}}}
  & Phi-3-mini-128k-instruct    & 3.8B \\
  & Phi-3-small-128k-instruct   & 7B \\
  & Phi-3-medium-128k-instruct  & 14B \\
\bottomrule[1pt]
\bottomrule[1.5pt]
\end{tabular}
\end{minipage}\hspace{4.5em}
\begin{minipage}[t]{0.4\textwidth}
\centering 
\begin{tabular}{@{} l l c @{}} 
\toprule[1.5pt]
\toprule[1pt]
\textbf{Model Family} & \textbf{Model ID} & \textbf{Size} \\
\addlinespace[2pt] \hdashline[1pt/1pt] \addlinespace[2pt]
\multirow{4}{*}{\shortstack[l]{\textbf{Qwen2.5}}}
  & Qwen2.5-7B-Instruct & 7B \\
  & Qwen2.5-14B-Instruct& 14B \\
  & Qwen2.5-32B-Instruct & 32B \\
  & Qwen2.5-72B-Instruct  & 72B \\
\addlinespace[2pt] \hdashline[1pt/1pt] \addlinespace[2pt]
\multirow{2}{*}{\shortstack[l]{\textbf{Qwen3}}}
  & Qwen3-8B   & 8B \\
  & Qwen3-14B   & 14B \\
\addlinespace[2pt] \hdashline[1pt/1pt] \addlinespace[2pt]
\multirow{1}{*}{\shortstack[l]{\textbf{Qwen2.5 Math}}}
  & Qwen2.5-Math-7B-Instruct   & 7B \\
\addlinespace[2pt] \hdashline[1pt/1pt] \addlinespace[2pt]
\multirow{2}{*}{\shortstack[l]{\textbf{Llama 3.1/3.3}}}
  & Llama-3.1-8B-Instruct   & 8B \\
  & Llama-3.3-70B-Instruct  & 70B \\
\addlinespace[2pt] \hdashline[1pt/1pt] \addlinespace[2pt]
\multirow{2}{*}{\shortstack[l]{\textbf{Gemma 2}}}
  & Gemma-2-9b-it   & 9B \\
  & Gemma-2-27b-it  & 27B \\
\addlinespace[2pt] \hdashline[1pt/1pt] \addlinespace[2pt]
\multirow{2}{*}{\shortstack[l]{\textbf{Mistral}}}
  & Mistral-7B-Instruct-v0.3    & 7B \\
  & Mistral-Nemo-Instruct-2407  & 12B \\
\addlinespace[2pt] \hdashline[1pt/1pt] \addlinespace[2pt]
\multirow{2}{*}{\shortstack[l]{\textbf{Deepseek}}}
  & DeepSeek-R1-Distill-Qwen-7B    & 7B \\
  & DeepSeek-R1  & 37B \\
\bottomrule[1pt]
\bottomrule[1.5pt]
\end{tabular}
\end{minipage}
\caption{The overview of teacher models we used in our experiments. For instruction tuning (left table), we directly use the teacher models defined in Magpie-Zoo dataset~\citep{stronger_are_not}. The right table presents the teacher models for math reasoning.}
\label{tab:teacher_models}
\end{table*}

\newpage
\onecolumn
\subsection{Implementation Details}
\label{appendix:implementation_details}

\paragraph{Training Setup.}
Table~\ref{tab:hyperparameters} and Table~\ref{tab:lora_hyperparameters} present the detailed hyper-parameters of full parameter fine-tuning and LoRA fine-tuning.\footnote{We follow the LoRA setting in these works~\citep{lora1,zhang2025guilo,lora2}}
We conduct our experiments on a server equipped with eight NVIDIA A100-SXM4-80GB GPUs.

\begin{table}[htbp]
\centering
\setlength\tabcolsep{4pt}
\fontsize{10}{11}\selectfont
\renewcommand{\arraystretch}{1.2}
\begin{tabular}{@{} l l @{}}
\toprule[1.5pt]\toprule[1.5pt]
\textbf{Hyper-parameter} & \textbf{Value} \\
\midrule
Learning Rate & $2 \times 10^{-5}$ \\
Number of Epochs & 2 \\
Number of Devices & 4 \\
Per-device Batch Size & 1 \\
Gradient Accumulation Steps & 8 \\
Effective Batch Size & 32 \\
Optimizer & AdamW \\
Learning Rate Scheduler & cosine \\
Warmup Steps & 100 \\
Max Sequence Length & 4096/16384 \\
\bottomrule[1.5pt]\bottomrule[1.5pt]
\end{tabular}
\caption{The hyper-parameters used for full parameter fine-tuning of models smaller than 14B. The max sequence length in the instruction tuning setting is 4,096, while that in the math reasoning setting is 16,384. Notably, the ``max\_position\_embeddings'' of Gemma-2 is 8,192; therefore, its max sequence length in the math reasoning setting is limited to 8,192.}
\label{tab:hyperparameters}
\end{table}

\begin{table}[htbp]
  \centering
  \setlength\tabcolsep{4pt}
\fontsize{10}{11}\selectfont
  \begin{tabular}{ll}
    \toprule[1.5pt]
    \toprule[1.5pt]
    \textbf{Hyper-parameter} & \textbf{Value} \\
    \midrule
    Learning Rate & \(1 \times 10^{-4}\) \\
    Number of Epochs & 2 \\
    Number of Devices & 4 \\
    Per-device Batch Size & 1 \\
    Gradient Accumulation Steps & 8 \\
Effective Batch Size & 32 \\
Optimizer & AdamW \\
    Lora Target & full \\
    Learning Rate Scheduler & cosine \\
    Warmup Ratio & 100 \\
    Max Sequence Length & 4096/16384 \\
    \bottomrule[1.5pt]\bottomrule[1.5pt]
  \end{tabular}
  \caption{The hyper-parameters used for LoRA fine-tuning of models larger than 14B. The max sequence length in the instruction tuning setting is 4,096, while that in the math reasoning setting is 16,384. Notably, the ``max\_position\_embeddings'' of Gemma-2 is 8,192; therefore, its max sequence length in the math reasoning setting is limited to 8,192.}
  \label{tab:lora_hyperparameters}
\end{table}

\begin{figure}[!th]
    \centering \centerline{\includegraphics[width=0.5\columnwidth]{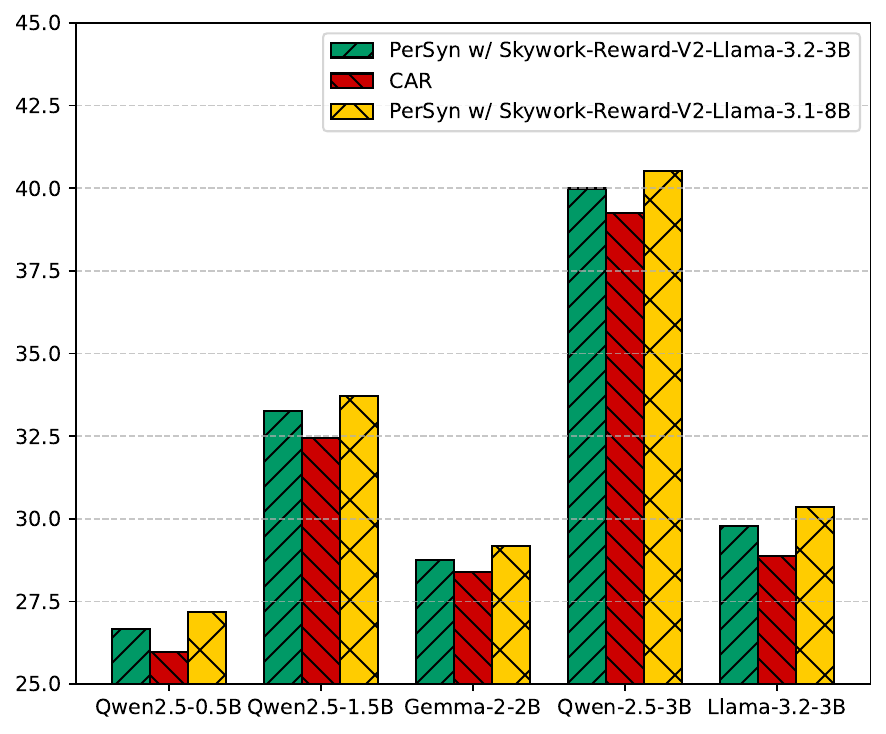}}
    \caption{The average results of the strong baseline \car and \textit{PerSyn} with strong and weak reward models in the instruction tuning setting.}
    \label{fig:different_reward_model}
\end{figure}

\paragraph{Reward Models.}
To study the impact of the reward model used in \S~\ref{sec:annotation_metric}, we conduct additional experiments in the instruction tuning setting, comparing the performance of \textit{PerSyn} with a weak reward model ( Skywork-Reward-V2-Llama-3.1-3B) and a strong reward model (Skywork-Reward-V2-Llama-3.1-8B).
The results in Fig.~\ref{fig:different_reward_model} show that using the weak reward model Skywork-Reward-V2-Llama-3.1-3B yields worse performance than the strong reward model Skywork-Reward-V2-Llama-3.1-8B, highlighting that \textit{PerSyn} benefits from a well-performing reward model. 
However, even with a weak reward model, \textit{PerSyn} still outperforms the strong baseline \car.
By default, we use Skywork-Reward-V2-Llama-3.1-8B reward model in all instruction tuning experiments.

\paragraph{Baselines Setup.}
We compare the following baselines to verify the effectiveness of our \textit{PerSyn} in our experiments: 
\textbf{1) Strong} refers to using the strongest single LLM as the teacher to synthesize data.
\textbf{2) Mix} indicates mix distillation~\citep{learnability_gap} that derives the synthetic dataset by mixing outputs from weak and strong teacher models.
In this work, we implement a variant of mix distillation by randomly assigning a teacher model's response to each prompt. Notably, the teacher models range from small to large sizes, thereby the synthetic dataset includes responses from both weak and strong teacher models.
\textbf{3) Family-Strong} represents choosing the strongest teacher model that is within the same family as the student model. 
\textbf{4) CAR} is a compatibility-adjusted reward metric proposed by~\citet{stronger_are_not} that selects a single teacher model based on the average performance of teacher model on the entire dataset.

We present the assigned teacher models of our compared baseline methods in the instruction tuning and math reasoning settings in 
Table~\ref{tab:baselines_instruction_tuning} and Table~\ref{tab:baselines_math_reasoning}.

\begin{table}[ht]
\centering
\label{tab:teacher_model_instruction_tuning}
\fontsize{9}{10}\selectfont
\setlength\tabcolsep{10pt}
\begin{tabular}{lccc}
\toprule[1.5pt]
\toprule[1pt]
\multirow{2}{*}{\textbf{Student Models}} & \multicolumn{3}{c}{\textbf{Instruction Tuning}} \\
\cmidrule(lr){2-4}
 & \textbf{Strong} & \textbf{Family-Strong} & \textbf{CAR} \\
\midrule
Qwen2.5-0.5B & Llama-3.1-405B-Instruct & Qwen2.5-72B-Instruct & Qwen2.5-3B-Instruct \\
Qwen2.5-1.5B & Llama-3.1-405B-Instruct & Qwen2.5-72B-Instruct & Qwen2.5-3B-Instruct \\
Gemma-2-2B   & Llama-3.1-405B-Instruct & Gemma-2-27b-it & Qwen2.5-72B-Instruct \\ 
Qwen2.5-3B   & Llama-3.1-405B-Instruct & Qwen2.5-72B-Instruct & Qwen2.5-3B-Instruct \\
Llama-3.2-3B & Llama-3.1-405B-Instruct & Llama-3.1-70B-Instruct & Qwen2.5-3B-Instruct \\
\bottomrule[1pt]
\bottomrule[1.5pt]
\end{tabular}
\caption{The assigned teacher models of our compared baseline methods in instruction tuning setting.}
\label{tab:baselines_instruction_tuning}
\end{table}

\begin{table}[ht]
\fontsize{9}{10}\selectfont
\setlength\tabcolsep{10pt}
\centering
\label{tab:teacher_model_math_reasoning}
\begin{tabular}{lccc}
\toprule[1.5pt]
\toprule[1pt]
\multirow{2}{*}{\textbf{Student Models}} & \multicolumn{3}{c}{\textbf{Math Reasoning}} \\
\cmidrule(lr){2-4}
 & \textbf{Strong} & \textbf{Family-Strong} & \textbf{CAR} \\
\midrule
Qwen2.5-0.5B & DeepSeek-R1-37B & Qwen2.5-72B-Instruct & Qwen2.5-Math-7B-Instruct \\
Qwen2.5-1.5B & DeepSeek-R1-37B & Qwen2.5-72B-Instruct & Qwen2.5-Math-7B-Instruct \\
Gemma-2-2B   & DeepSeek-R1-37B & Gemma-2-27b-it & Llama-3.1-8B-Instruct \\ 
Qwen2.5-3B   & DeepSeek-R1-37B & Qwen2.5-72B-Instruct & Qwen2.5-Math-7B-Instruct \\
Llama-3.2-3B & DeepSeek-R1-37B & Llama-3.3-70B-Instruct & Qwen2.5-Math-7B-Instruct \\
\bottomrule[1pt]
\bottomrule[1.5pt]
\end{tabular}
\caption{The assigned teacher models of our compared baseline methods in math reasoning setting.}
\label{tab:baselines_math_reasoning}
\end{table}

\end{document}